\documentclass[sigconf]{acmart}
\acmConference[AST 2024]{5th International Conference on Automation of Software Test}{April 2024}{Lisbon, Portugal}
\usepackage{tabularx}
\DeclareMathOperator*{\ARGMAX}{argmax}
\def\mathbi#1{\textbf{\em #1}}

\usepackage{paralist}

\usepackage{subcaption}
\usepackage{graphicx}
\AtBeginDocument{%
  }

\copyrightyear{2024} 
\acmYear{2024} 
\setcopyright{rightsretained} 
\acmConference[AST '24]{5th ACM/IEEE International Conference on Automation of Software Test (AST 2024)}{April 15--16, 2024}{Lisbon, Portugal}
\acmBooktitle{5th ACM/IEEE International Conference on Automation of Software Test (AST 2024) (AST '24), April 15--16, 2024, Lisbon, Portugal}\acmDOI{10.1145/3644032.3644453}
\acmISBN{979-8-4007-0588-5/24/04}

\begin{document}

\title[Properties affecting transferability of adversarial attacks on quantized networks]{Properties that allow or prohibit transferability of adversarial attacks among quantized networks}

\author{Abhishek Shrestha}
\email{abhishek.shrestha@fokus.fraunhofer.de}
\author{Jürgen Großmann}
\email{juergen.grossmann@fokus.fraunhofer.de}
\affiliation{%
  \institution{Fraunhofer-Institut für Offene Kommunikationssysteme FOKUS}
  \streetaddress{Kaiserin-Augusta-Allee 31}
  \city{Berlin}
  %\state{Berlin}
  \country{Germany}
}

%\renewcommand{\shortauthors}{Abhishek and Jürgen}

\begin{abstract}
  Deep Neural Networks (DNNs) are known to be vulnerable to adversarial examples. Further, these adversarial examples are found to be transferable from the source network in which they are crafted to a black-box target network. As the trend of using deep learning on embedded devices grows, it becomes relevant to study the transferability properties of adversarial examples among compressed networks. In this paper, we consider quantization as a network compression technique and evaluate the performance of transfer-based attacks when the source and target networks are quantized at different bitwidths. We explore how algorithm specific properties affect transferability by considering various adversarial example generation algorithms. Furthermore, we examine transferability in a more realistic scenario where the source and target networks may differ in bitwidth and other model-related properties like capacity and architecture. We find that although quantization reduces transferability, certain attack types demonstrate an ability to enhance it. Additionally, the average transferability of adversarial examples among quantized versions of a network can be used to estimate the transferability to quantized target networks with varying capacity and architecture.
\end{abstract}

\begin{CCSXML}
<ccs2012>
   <concept>
       <concept_id>10002978.10003022</concept_id>
       <concept_desc>Security and privacy~Software and application security</concept_desc>
       <concept_significance>500</concept_significance>
       </concept>
 </ccs2012>
\end{CCSXML}

\ccsdesc[500]{Security and privacy~Software and application security}

\keywords{Deep neural networks, Quantization, Adversarial examples, Transferability}

%\received{20 November 2023}
%\received[revised]{12 December 2023}
%\received[accepted]{12 December 2023}

\maketitle

\section{Introduction}
The last decade has seen deep learning become increasingly popular in multitude of domains like healthcare \cite{intro_medical_imaging}, speech-recognition \cite{intro_speech_recog}, self-driving cars \cite{intro_autonomous_veh}, and collision avoidance systems \cite{intro_collision_1, intro_collision_2}. One of the prominent domains where deep learning is increasingly being used is the embedded systems. For instance, IOT devices use AI for data collection and processing \cite{ai_iot}; mobile devices use them for functionalities like face recognition \cite{intro_face_recog}, voice assistants \cite{intro_voice_recog}, hardware emulation in cameras \cite{intro_hardware_emu}, and so forth.

However, the evolution of AI as a powerful technology to solve complicated tasks has been, in part, made possible due to the rise of powerful hardware capable of running large DNN models \cite{impact_low_bit}. Given that the embedded devices are inherently limited on resources, implementation of deep learning on the device itself often involves optimization of the base models. Model optimization is a topic of active research and several methods have been developed to fit the purpose \cite{pruning_paper, quantizing_dnn, mobilenets}. One of such methods is to compress a model via quantization \cite{survey_qualtization_tech}. This involves reducing the computational complexity and size of a model by reducing the precision of the network parameters from default float32 to smaller bitwidths.

DNNs are found to be vulnerable to data samples with deliberately added and often imperceptible perturbations \cite{intriguing_properties}. Benign images that are otherwise classified correctly by a network, when subjected to these perturbation vectors, can cause classifiers to misclassify the perturbed images at a high rate \cite{exp_harness, space_for_adv}. These perturbed images or \emph{adversarial examples} are a severe threat to the usability of DNNs in safety-critical domains as they can effectively fool a network into making wrong decisions inspired by an adversary.

Moreover, adversarial examples are observed to be transferable. Examples generated on one classifier are found to be effective on other classifiers trained to perform the same task \cite{space_for_adv, understand_enhance, exp_harness, physical_world}. This enables an adversary to mount a black-box attack on a target network with adversarial images crafted in another network. Here, the network in which the adversarial examples are created is termed as the \emph{source network} and the network in which the adversarial attacks are applied is called the \emph{target network}.

In the context of embedded systems, adversarial transferability has even more severe consequences due to the growing ubiquity of these devices. For instance, an attacker can gain access to a model deployed in an edge device and use it to craft adversarial examples; the transferability property of adversarial examples would then enable the attacker to perform transfer-based black-box attacks on a secure base model. Moreover, the attack can be transferred to other compressed models derived from the same base model \cite{to_compress}. Conversely, a manufacturer can use a publicly available base model and roll-out compressed versions of it on his devices. Attackers can then use the base model to craft attacks against the private devices. Thus, it becomes crucial to understand the factors that affect the transferability of adversarial attacks across compressed networks. 

The main objective of this paper is to analyse the conditions and properties that facilitate or hinder transferability of adversarial attacks among networks of different bitwidths. This involves investigation of how various adversarial attack generation algorithms and model-specific properties affect the transferability of adversarial examples when source and target networks differ in quantization levels. More specifically, the goal is to answer the following research questions:

\begin{itemize}\label{intro:RQ}

	\item \textbf{RQ1}: What causes some algorithms to have better or worse transferability across variably quantized systems? 
	
	\item \textbf{RQ2}: How do model-related properties like model architecture and capacity affect the transferability of attacks between networks of different bitwidths?
	
\end {itemize}

To answer these questions we perform a comprehensive study of transferability among variable bitwidth networks. The main contributions of this paper are:

\begin{itemize}
\item By considering a broad range of algorithms when creating adversarial examples, we evaluate the performance of quantized and full-precision networks when the attacks are transferred from another network having a different bitwidth. 

Our analysis of transferability properties of the Universal Adversarial Perturbation (UAP) attack \cite{uap_paper} reveals that it allows higher magnitudes of perturbation than the internal algorithm used to craft the UAP while still keeping the adversarial image recognizable to human observers. This increases the effectiveness of UAPs at higher distortions which suggests that the performance of an attack may be improved by aggregating the individual perturbations from multiple images to create a UAP. Further, the observed poor transferability of Jacobian Saliency Map Based Attack (JSMA) \cite{jsma_paper} indicates that the attacks that depend on individual feature modifications may have poor transferability. Moreover, search-based attacks like the Boundary Attack may have poor transferability due to direction of perturbation introduced by the algorithm.

\item We consider a more realistic scenario where the target model could be different from source in terms of model capacity or architecture or both. In such scenarios, we show that the overall success-rate of a transfer-based attack can be estimated by observing the performance of the attack when it is transferred among different bitwidth versions of the source model.

\end{itemize}

The rest of the paper is organised as follows: Section \ref{sec:related_works} mentions prior studies on transferability of adversarial attacks. Section \ref{sec:background} provides a short background on network quantization, adversarial examples, and attack generation methods used in this work. Section \ref{sec:experiments} presents the experiments together with the evaluation of the results. Section \ref{sec:discussion} reflects on the results. Finally, we conclude in Section \ref{sec:conclusion}.

\section{Related Works} \label{sec:related_works}
There are already existing bodies of work that make considerable contributions in postulating how different factors might affect transferability of adversarial examples. Findings in \cite{delving_trans} indicate that the attack algorithm used to craft the adversarial examples can affect their transferability. Authors find that some fast gradient-based attacks like the Fast Gradient Sign Method (FGSM) \cite{exp_harness} are less transferable than search-based attacks. Further, observations in \cite{understand_enhance} show that factors such as model capacity, accuracy, and architecture of both source and target models affect transferability. The authors additionally note that a smooth loss surface on the source network results in better chances of alignment of source and target gradients, resulting in better transferability. In \cite{bugs_not_features}, authors offer a new perspective on transferability as they argue that it is the property of the data distribution itself and that highly accurate models trained with non-robust or brittle features are more vulnerable to transfer-based attacks. 

Thus, it is clear that there are multiple factors at play when considering transferability of adversarial attacks. Nevertheless, the dynamics of these factors change when considering transferability among networks with varying bitwidths. For instance, \cite{understand_enhance} and \cite{space_for_adv} state that high accuracy models have similar decision boundaries, and thus transferability is high among such networks. However, the transferability of attacks between a full-precision and its quantized version is observed to be considerably low even when both of the networks have similar accuracy \cite{impact_low_bit}. 

Prior works on transferability of adversarial examples in regard to quantized networks address how various algorithms can affect transferability \cite{uap_transfer, impact_low_bit, to_compress}; however, the focus is mainly on attacks that either use loss gradients or estimate them to craft adversarial attacks. Moreover, the current state of the art mostly consider cases where both source and target networks are different bitwidth versions of the same network, but in real-world settings, an attacker may not have knowledge of the quantization level and other model-specific properties of the target network. Our objectives with this study is to address these gaps.

\section{Background}\label{sec:background}
\subsection{Deep Neural networks (DNNs)} \label{sec:dnn}
%A DNN can be defined as a function that maps a high-dimensional input to a vector output. 
Given an input $\mathbf{x} \in \mathbb{R}^{m}$ and the corresponding output vector $\mathbf{y} \in \mathbb{R}^{n}$, a DNN is a classification function $f$: $\mathbf{x} \mapsto \mathbf{y}$, where $m$ is high-dimensional. In addition to the input $\mathbf{x}$, the output $\mathbf{y}$ also depends on the network parameters $\theta$ which the network learns during the training process. 

For the n-class classifiers considered in this work, the output vector $\mathbf{y}$ is a probability distribution, that is: ${y_1 + y_2 + y_3 + \ldots + y_n = 1}$ and ${0 \leq ({y_i})_{i=1}^{n} \leq 1}$. Each element ${y}_{i}$ in $\mathbf{y}$ represents the probability that the input $\mathbf{x}$ has class $i$. Thus, for a network $\mathbf{y} = f(\mathbf{x})$, the label $y$ assigned by the classifier to input ${\mathbf{x}}$ is:

\begin{equation} \label{net_label}
	\ARGMAX_i f_{i}(\mathbf{x}) = y
\end{equation}

Here, $y_i = f_i(\mathbf{x})$ is the $i^{th}$ output of the network.

\subsection{Adversarial Examples} \label{sec:adv_examples}
Adversarial examples are created by adding computed perturbations to a clean image. The resulting distorted samples look similar to their original counterparts and are still classified correctly by human oracles, but the perturbations are enough for a classifier to change the output class probabilities, possibly leading to a misclassification \cite{huang_survey, intriguing_properties, exp_harness, physical_world}.

Let ${y^{true}}$ be the true label of a clean sample $\mathbf{x}$, then from Equation \ref{net_label} we have: ${\ARGMAX_i f_{i}(\mathbf{x}) = y^{true}}$. If $\mathbf{x}$ is subjected to a perturbation vector $\eta \in \mathbb{R}^{m}$, resulting in a perturbed example ${\mathbf{x}^{adv}}$ that causes misclassification, then:

\begin{equation} \label{adv_ex}
	\ARGMAX_i f_{i}(\mathbf{x}^{adv}) \neq y^{true}
\end{equation}

\subsection{Crafting Algorithms}\label{sec:craft_adv}

The perturbation introduced to create an adversarial example is not random. In fact, it is much harder to cause misclassification by adding perturbations in random directions \cite{space_for_adv}. An adversarial example generation algorithm adds perturbations to the original image in a specific direction, computed within the steps of the algorithm.

\textbf{Fast Gradient Sign Method (FGSM)}  \label{sec:alg:FGSM}
 \cite{exp_harness} is one of the simplest methods to craft adversarial examples. The single-step algorithm adds max-norm constrained perturbation in the direction of the loss gradient of the network to create adversarial samples.

\begin{equation} \label{fgsm_attack}
\begin{aligned}
	\mathbf{x}^{adv} = \mathbf{x} + \varepsilon sign(\nabla J_{\mathbf{x}}(\mathbf{x}, y, \theta)) 	
\end{aligned}	
\end{equation}

Here, $\nabla J_{\mathbf{x}}$ is the gradient of the loss function ${J(\mathbf{x},y,\theta)}$, computed with respect to input $\mathbf{x}$. Since loss gradient gives the direction of the largest increase in loss, adding a small perturbation aligned with this direction can cause significant increase in loss at high dimensions. $\varepsilon$ is the $L_\infty$ norm of the perturbation which also gives the distance between $\mathbf{x}$ and $\mathbf{x}^{adv}$.

\textbf{Jacobian Saliency Map based Attack (JSMA)}  \label{sec:alg:JSMA}
\cite{jsma_paper} computes adversarial examples by creating a direct mapping between input and output variations and using this map to isolate features that are most effective in causing classification change. The algorithm uses forward derivative of the function learned by the network (given by Equation \ref{jsma_attack}) to construct a saliency map which filters the features that are most important based on the the given criteria. Equation \ref{salency_map_criteria} provides a basic filter criteria as defined in \cite{jsma_paper}.

\begin{equation} \label{jsma_attack}
\begin{aligned}
	\nabla f(\mathbf{x}) = \frac{\partial f(\mathbf{x})}{\partial \mathbf{x}} = \left[\frac{\partial f_j(\mathbf{x})}{\partial x_i}\right]_{i\in 1, \ldots ,m, j \in 1, \ldots ,n}	
\end{aligned}	
\end{equation}

\begin{equation} \label{salency_map_criteria}
\begin{aligned}
	S(\nabla f(\mathbf{x}), t)[i] = \left \{ \begin{array}{ll}
	0 \mbox{~if~} \frac{\partial f_t(\mathbf{x})}{\partial {x}_i} < 0  \mbox{~or~}  \sum_{j \neq t} \frac{\partial f_j(\mathbf{x})}{\partial {x}_i} > 0 \\ \\
	\left( \frac{\partial f_t(\mathbf{x})}{\partial {x}_i} \right)  \left |\sum_{j \neq t} \frac{\partial f_j(\mathbf{x})}{\partial {x}_i}\right |  \mbox{~otherwise~}  
 	\end{array}\right.
\end{aligned}	
\end{equation}

In Equation \ref{salency_map_criteria}, t is the target class to which the input is to be misclassified, that is, $t \neq y^{true}$. $S(\nabla f(\mathbf{x}), t)[i]$ is the saliency map computed for $i^{th}$ feature. 

Thus, as per Equation \ref{salency_map_criteria}, from the Jacobian matrix (given by Equation \ref{jsma_attack}), features that increase the target class probability and at the same time decrease the probabilities of all other classes are weighed and the feature with the highest value is selected. In each iteration, selected feature is perturbed by a defined amount. This is continued till adversarial goal of misclassification, that is, ${\ARGMAX_i f_{i}(\mathbf{x}) = t}$, is reached. 

\textbf{Universal Adversarial Perturbation (UAP) Attack}  \label{sec:alg:UAP}
\cite{uap_paper} is quite different from other algorithms because it does not create adversarial sample but instead results in a single perturbation vector that can be added to any point in the dataset to create adversarial examples. Such perturbation vector is called Universal Adversarial Perturbation or UAP. 

If $X = \{\mathbf{x}_1, \mathbf{x}_2, \ldots ,\mathbf{x}_n\}$ be a dataset sampled from a data distribution $\mu$, then the basic idea is to sequentially iterate though each image in $X$ and compute a perturbation vector $\Delta \mathbi{v}_k$ that sends the current image $\mathbf{x}_k + \mathbi{v}$ across the decision boundary of the network. Every iteration also updates the overall perturbation  $\mathbi{v}$ with $\Delta \mathbi{v}_k$. The algorithm stops when the computed perturbation is able to fool a pre-defined number of images in $X$. The perturbation vector $\Delta \mathbi{v}_k$ can be computed using any algorithm, in the case of this paper, we use FGSM.

\textbf{The Boundary Attack (BA)} \cite{ba_paper} uses model's decisions on the input points to craft adversarial examples. The attack is conceptually simple. Given a benign image, the algorithm initializes a random adversarial image. The initialized random image is then subjected to multiple perturbations iteratively. Each perturbation decreases the $L_2$ distance between the benign and the adversarial image. When adding these perturbations, the algorithm queries the model with the perturbed image as input to make sure that the perturbation still keeps the image outside the decision boundary of the true class. Thus, the initial random image is moved along the decision boundary between the adversarial and non-adversarial region until minimum $L_2$ distance between the benign and the adversarial image is achieved. At the end of the run, the initial image remains adversarial but looks like the benign image to human observers. 

The boundary attack does not require access to model parameters and architecture, thus it can operate under \emph{black-box} scenarios. Further, it does not use gradients to create adversarial examples.

\textbf{The Carlini-Wagner (CW) Attack} \cite{cw_attack} is one of the most powerful adversarial attacks. It uses an optimization function that iteratively performs gradient descent towards the target adversarial class while keeping the distance between original and the adversarial example minimal. Equation \ref{cw_attack} defines the optimization problem the attack algorithm solves.

\begin{equation} \label{cw_attack}
\begin{aligned}
	&\mbox{~minimize~} \| \mathbf{x}^{adv} - \mathbf{x}\|_2^2 + c \cdot l(\mathbf{x}^{adv}); 
\mbox{~where,~} & \\
	l(\mathbf{x}^{adv}) &= \max \left(Z_i(\mathbf{x}^{adv}) - \max \left(Z_t(\mathbf{x}^{adv}) : t \neq i\right) + \kappa, 0 \right)
\end{aligned}
\end{equation}

In the above equation, $Z_{i}(\mathbf{x}^{adv})$ is the true label while $Z_{t}(\mathbf{x}^{adv})$ is any label other than ${i}$, thus the loss function $l$ ensures misclassification. Further, it can be seen that the attack uses logits $Z(\mathbf{x})$ instead of output from the softmax layer. This enables the attack to bypass defensive techniques such as defensive distillation \cite{def_distill} which work by smoothening the model's output. Similarly, $c$ is a non-negative constant that balances misclassification confidence and distance between adversarial and original sample. It is determined during run-time using binary search. The attack also allows to define misclassification confidence $\kappa$. Increasing the misclassification confidence increases the probability that the attack becomes more transferable but at the cost of higher distortions. 

As can be seen, the equation focuses on only $L_2$ metric when creating attacks. In their paper, authors also define $L_0$ and $L_\infty$ variants; however, our analysis only includes $L_2$ norm as it is the strongest and the attack was originally formulated for this norm, while other norms were adapted from $L_2$.

\subsection{Quantization as a model optimization technique.}

Quantization reduces the bitwidth of network values such that computational complexity during training and inference is reduced \cite{quantizing_dnn}. By using lower bitwidth numbers rather than default float32 values, floating-point multiplications during convolution operations can be replaced by bitwise operations which are much faster.

In this paper, network quantization is performed during training using a method called DoReFa-Net \cite{dorefa_net_paper}. For any real number $r_i \in [0,1]$, the corresponding n-bit quantized output value $r_o$ is computed during forward pass using the quantization function: $r_o = \frac{1}{2^n-1} \mbox{round}((2^n-1) \mbox{~~} r_i)$. However, since a continuous function with a small finite range always has zero gradient with respect to its input \cite{dorefa_net_paper}, the quantization function itself is not differentiable. This creates a problem during back-propagation. For instance, if $c$ is the cost function, then during backward pass, computation as in Equation \ref{undefined_grad} requires $\frac{\partial r_o}{\partial r_i}$ which is not defined.

\begin{equation} \label{undefined_grad}
\begin{aligned}
\frac{\partial c}{\partial r_i} = \frac{\partial c}{\partial r_o} \cdot \frac{\partial r_o}{\partial r_i}
\end{aligned}
\end{equation}

One of the solution to this problem is to estimate the value of $\frac{\partial r_o}{\partial r_i}$, given that $\frac{\partial c}{\partial r_o}$ is properly defined. The estimators that allow defining custom $\frac{\partial r_o}{\partial r_i}$ are called Straight Through Estimators or STEs \cite{ste_paper}. DoReFa-Net uses $\frac{\partial c}{\partial r_i} = \frac{\partial c}{\partial r_o}$ as STE during activation and weight quantization.

\section{Experiments} \label{sec:experiments}
We use CIFAR10 \cite{cifar_dataset} and MNIST \cite{mnist_dataset} datasets in all experiments. For MNIST dataset we trained a simple 8-layer CNN (called Mnist A for easier reference), while for CIFAR10, a ResNet20 network was trained. The architectures of both CIFAR10 and MNIST models are defined using the examples from the Tensorpack repository \cite{tensorpack}. For both MNIST and CIFAR10 models, six quantization bitwidths are considered: 1, 2, 4, 8, 12, and 16 bits. Table~\ref{tbl:Models} shows accuracies of all quantized and full-precision models. Training and architecture details are included in Appendix \ref{app:model_details}.

\begin{table}[htpb]
\caption[Test set accuracy of the full-precision (FP) and quantized versions of the MNIST and CIFAR10 models.]{Test set accuracy of the full-precision (FP) and quantized versions of the MNIST and CIFAR10 models.}
\Description{Test set accuracy of the full-precision (FP) and quantized versions of the MNIST and CIFAR10 models.}
\label{tbl:Models}
\centering
	\begin{tabular}	{m{0.072\textwidth}  m{0.034\textwidth}  m{0.034\textwidth} m{0.034\textwidth}  m{0.034\textwidth} m{0.034\textwidth}  m{0.043\textwidth}  m{0.043\textwidth}}	
		\toprule
		\textbf{Model ID}& \textbf{FP} & \textbf{1 bit} & \textbf{2 bit} & \textbf{4 bit} & \textbf{8 bit} & \textbf{12 bit} & \textbf{16 bit}\\ 
		\toprule	
		%-------------------------------
		Mnist A  & 0.991  & 0.991  & 0.991 & 0.992  & 0.992 & 0.991  & 0.990\\ 	
		\midrule
		Resnet20 & 0.891  & 0.834  & 0.865 & 0.847  & 0.829 & 0.843  & 0.842\\ 
		\bottomrule
	\end{tabular}	
\end{table}

Transferability properties of the five attack algorithms mentioned in Section~\ref{sec:craft_adv} were studied. For each attack type, different sets of hyperparameter values were used for MNIST and CIFAR10 models. These values are as depicted in Table~\ref{tbl:mnisthyperparam}. As seen in the table, $\varepsilon$ for FGSM is the magnitude of distortion added to the original image. For JSMA, $\gamma$ and $\theta$ represent the total percentage of pixels allowed to be distorted and distortion added per pixel (per iteration), respectively. For UAP attack, FGSM was used internally to generate UAP (see Section~\ref{sec:craft_adv}); $\varepsilon$ thus represents the perturbation magnitude for FGSM, while $\xi$ represents the maximum magnitude of distortion allowed, measured in $L_\infty$ norm as recommended in \cite{uap_transfer}. For BA, \emph{i} represents the number of iterations. Similarly, for CW attack, $\kappa\geq0$ is the confidence parameter, while \emph{i} is the maximum number of iterations (of gradient descent) per image, $b_{s}$ is the binary search steps to determine proper $c$, and $c_i$ is the initial value of $c$\footnote{The values of $c_i$ and $b_{s}$ were selected as per the usages in the original paper, $\kappa$ was selected such that images were clearly distorted but still recognizable.}.

\begin{table}[htpb]
\caption{Attack hyperparameter values for the full-precision (FP) and quantized versions of the MNIST and CIFAR10 models.}
\Description{Attack hyperparameter values for the full-precision (FP) and quantized versions of the MNIST and CIFAR10 models.}
\label{tbl:mnisthyperparam}
\centering
	\begin{tabular} 	
	{m{0.073\textwidth} | m{0.041\textwidth} | m{0.02\textwidth} m{0.034\textwidth} | m{0.02\textwidth} m{0.02\textwidth} | m{0.02\textwidth} | m{0.002\textwidth}m{0.009\textwidth}m{0.009\textwidth}m{0.009\textwidth}}		
		\toprule
		& \multicolumn{10}{c}{\textbf{Hyperparameter values}}\\
		\midrule
		&\textbf{FGSM} & \multicolumn{2}{c|}{\textbf{JSMA}} & \multicolumn{2}{c|}{\textbf{UAP}} & \textbf{BA} & \multicolumn{3}{c}{\textbf{CW}}\\
		\textbf{Model ID} & $\varepsilon$ & $\theta$ & $\gamma$ (\%) & $\varepsilon$ & $\xi$  & \emph{i} & $\kappa$ & \emph{i} & $b_{s}$ & $c_{i}$\\
		\toprule	
		%-------------------------------
		%-------------------------------
		Mnist A  &0.25  	&1   	&10 	&0.1 	&0.6 	&15	&5	&25  &20 &0.01\\ 	
		Resnet20  &0.05   	&0.3    &5 		&0.01 	&0.1  	&12		&5  &25  &20 &0.01\\ 	
		\bottomrule
	\end{tabular}
\end{table}

These hyperparameter values were selected such that the resulting adversarial images are distorted but yet recognizable to human observers. Thus, for all attack types, we consider upper limit of transferability by creating adversarial samples that are as distorted as possible. Different values of hyperparameters were selected for the same attack type for MNIST and CIFAR10 because for natural datasets like CIFAR10, relatively small distortion is enough to highly distort the image \cite{uap_transfer}.

Each of these attacks were implemented using the Adversarial Robustness Toolbox (ART) library \cite{art_paper}. Regarding quantization, we quantize both weights and activations (and not gradients).

Akin to \cite{impact_low_bit}, we measure the effectiveness of an attack on a network in terms of adversarial accuracy. This metric gives the network's accuracy against adversarial examples and is computed as the ratio of the number of adversarial samples classified correctly by the network to the total number of samples. Thus, a higher adversarial accuracy means the network is more robust against adversarial samples.

\subsection{Transferability of adversarial examples}
The objective of this experiment was to evaluate how algorithm specific properties affect transferability. For both CIFAR10 and MNIST models, 1000 randomly selected clean samples that were classified correctly by both source and target models were used to create adversarial samples. These were then transferred among different bitwidth versions of the same model and the resulting adversarial accuracy of the target model was measured. This process was repeated 3 times and the average adversarial accuracy was noted.
%, so each cell in Figure \ref{fig:transferability_res_resnet20} represents average of three separate runs. 

\begin{figure*}[htpb]
%hp = put figure "here" in a page for "only float"
	\centering
	%---FGSM_005------
	\begin{subfigure}{.34\textwidth}
	  \centering
	  \includegraphics[width=0.99\linewidth]{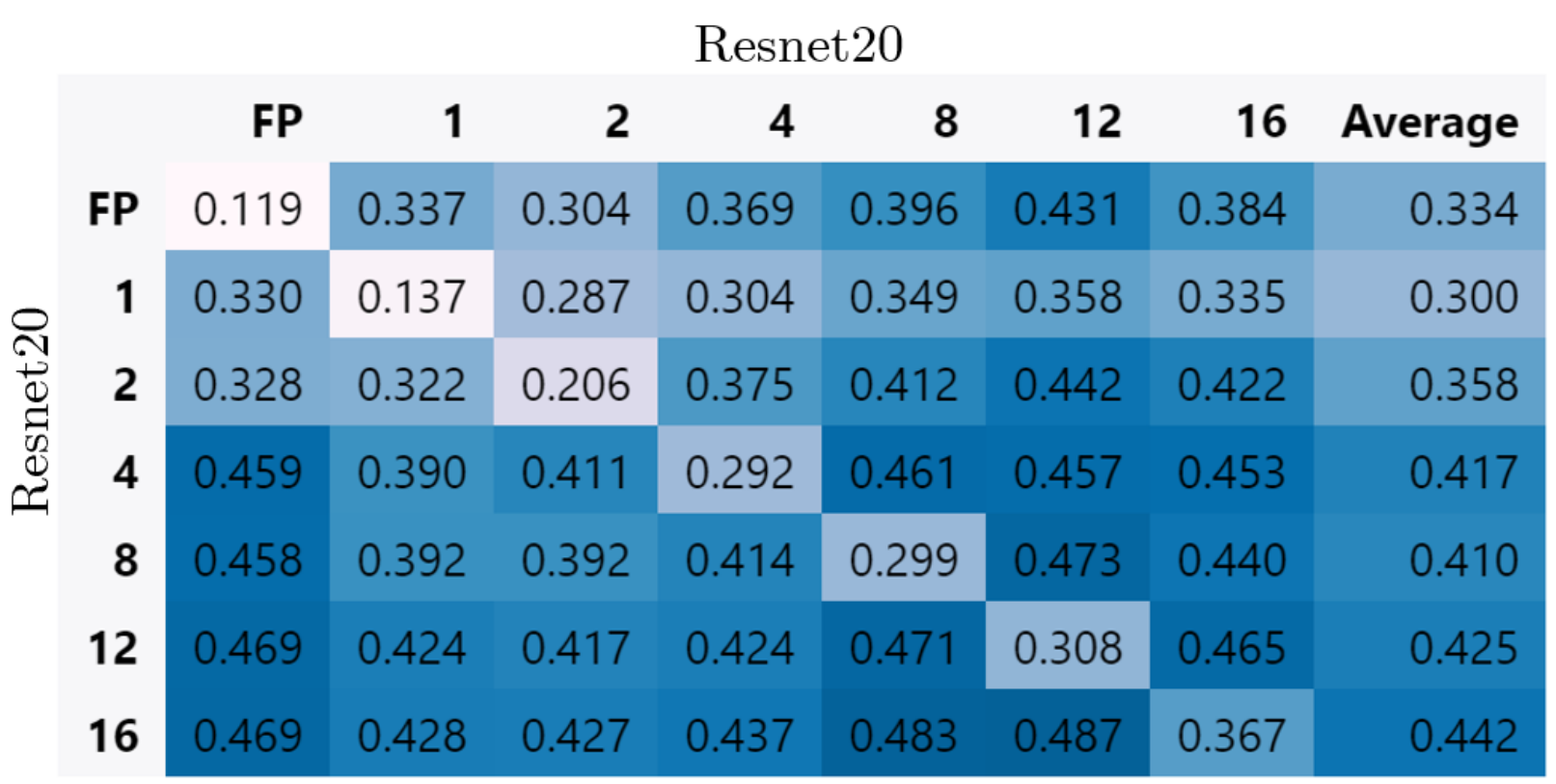}
	  \caption*{FGSM ($\varepsilon = 0.05$)}
	  \label{fig:cifar_FGSM_2}  
	\end{subfigure}%
	%---JSMA_T_03_G_005------
	\begin{subfigure}{.34\textwidth}
	  \centering
	  \includegraphics[width=0.99\linewidth]{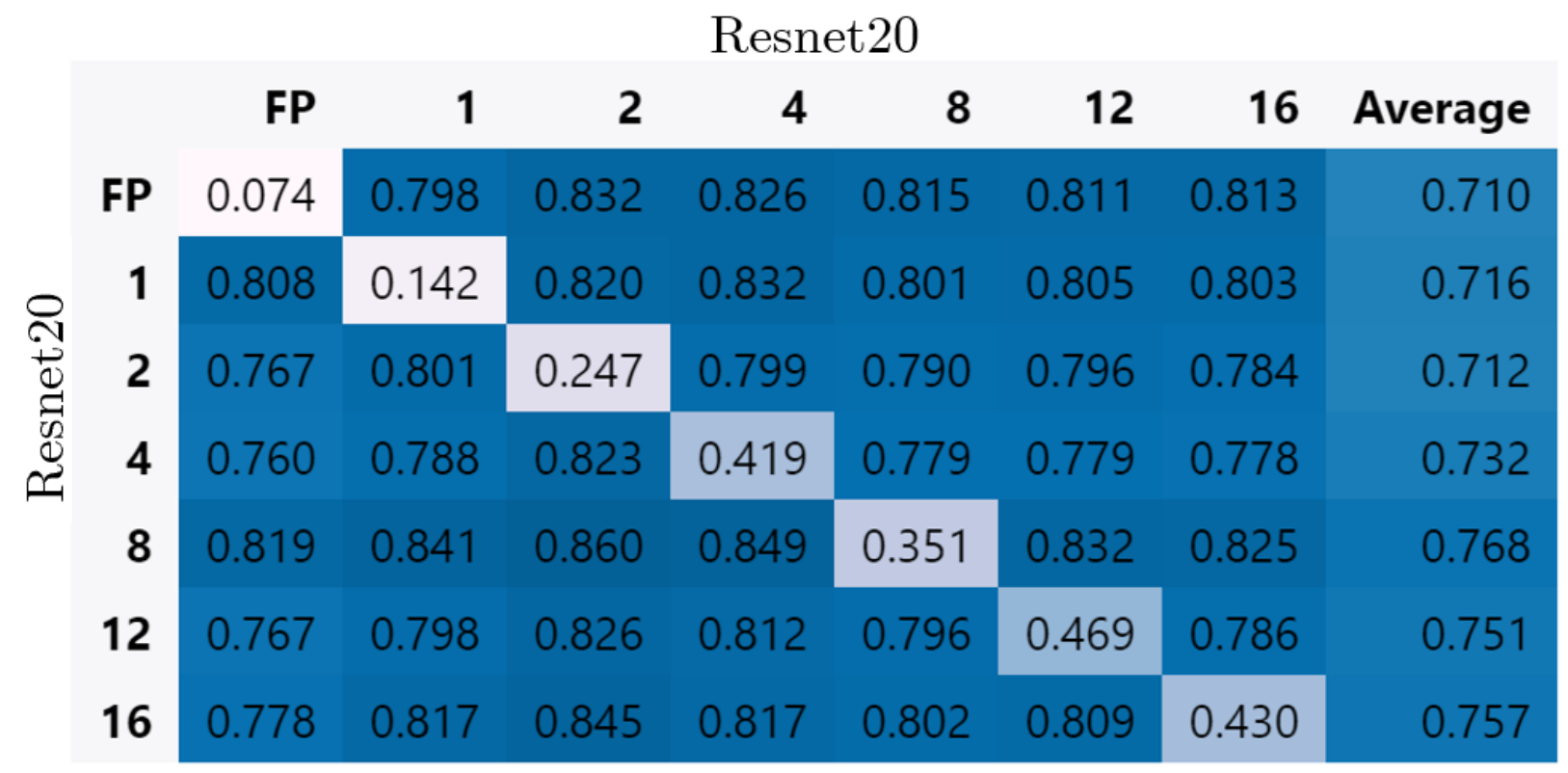}
	  \caption*{JSMA ($\theta = 0.3$, $\gamma = 5\%$)}
	  \label{fig:cifar_JSMA_2}
	\end{subfigure}%
	%---UAP_EP_001_XI_01------
	\bigskip
	\begin{subfigure}{.34\textwidth}
	  \centering
	  \includegraphics[width=0.99\linewidth]{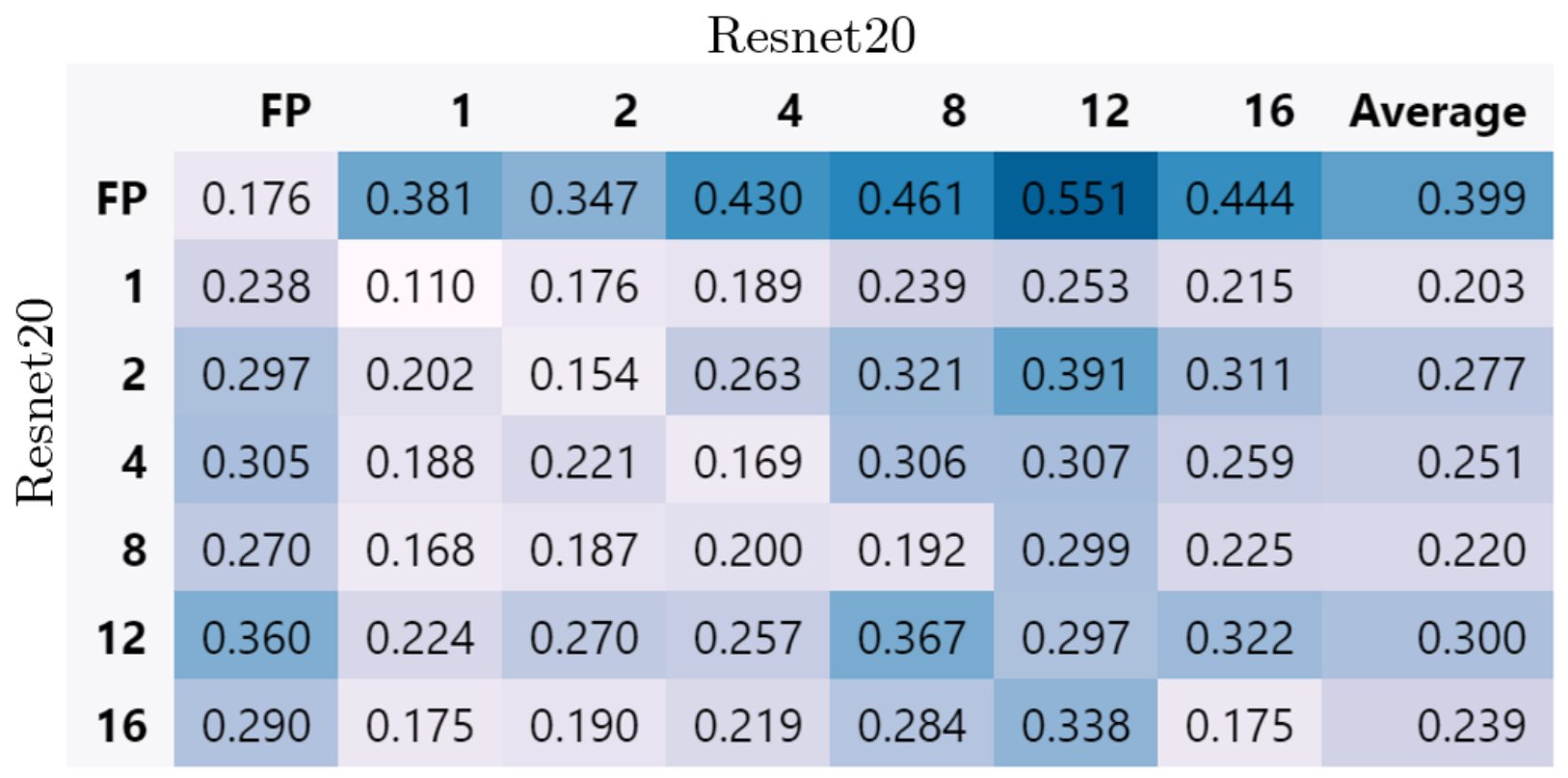}
	  \caption*{UAP ($\varepsilon = 0.01$, $\xi = 0.1$)}
	  \label{fig:cifar_UAP_2}
	\end{subfigure}
	%---CW_BSS_20_K_25_ITT_10------
	\begin{subfigure}{.34\textwidth}
	  \centering
	  \includegraphics[width=0.99\linewidth]{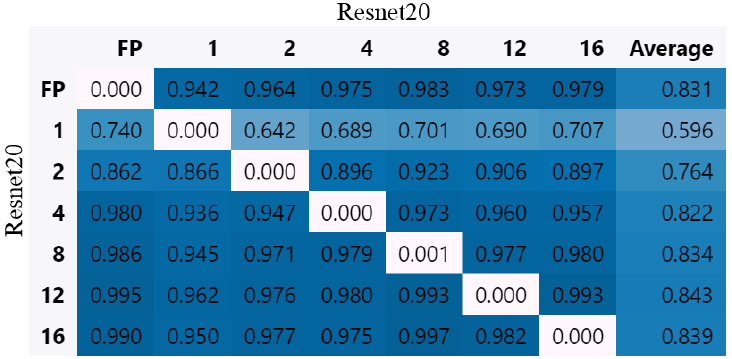}
	  \caption*{CW ($\kappa = 5$, $i = 25$, $b_{s} = 20$)}
	  \label{fig:cifar_CW}
	\end{subfigure}\hspace{2.5mm}%
	%---BA_12_ITT------	
	\begin{subfigure}{.34\textwidth}
	  \centering
	  \includegraphics[width=0.99\linewidth]{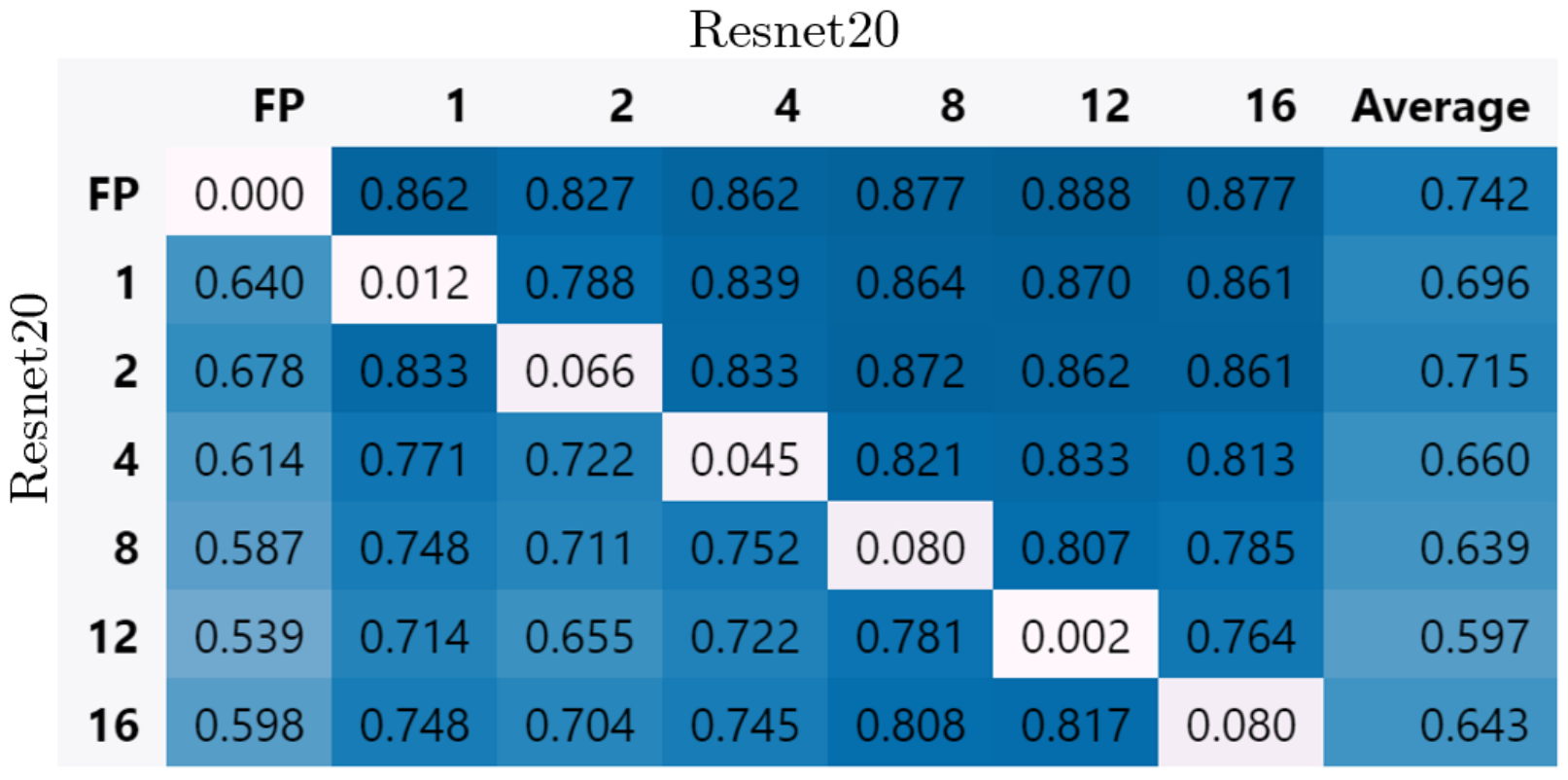}
	  \caption*{BA ($i = 12$)}
	  \label{fig:cifar_BA_1} 
	\end{subfigure}
	\captionsetup{singlelinecheck=off}
	\caption[Transferability of adversarial attacks among different bitwidth versions of the Resnet20 model.]{Transferability of adversarial attacks among different bitwidth versions of the Resnet20 model. In each matrix, rows indicate the source networks, while columns indicate the corresponding target networks. Row and column headers specify the bitwidth of the source and target models, respectively. The source and target model IDs are labelled alongside the corresponding headers. Cell values correspond to the adversarial accuracy of the target. Higher values (darker colours) indicate less transferability, while lower values (lighter colours) indicate more transferability. The diagonal values correspond to attack performance on the source (the source and target are the same model). Each value in the "Average" column indicates the average adversarial accuracy of all target models (one complete row) against a single attack source.}
	\Description{Transferability of adversarial attacks among different bitwidth versions of the Resnet20 model. In each matrix, rows indicate the source networks, while columns indicate the corresponding target networks. Row and column headers specify the bitwidth of the source and target models, respectively. The source and target model IDs are labelled alongside the corresponding headers. Cell values correspond to the adversarial accuracy of the target. Higher values (darker colours) indicate less transferability, while lower values (lighter colours) indicate more transferability. The diagonal values correspond to attack performance on the source (the source and target are the same model). Each value in the "Average" column indicates the average adversarial accuracy of all target models (one complete row) against a single attack source.}
	\label{fig:transferability_res_resnet20}
\end{figure*}

Figure \ref{fig:transferability_res_resnet20} reports the transferability results for Resnet20 models. Results for MNIST models are similar and are excluded for brevity.

%-------------------------------------------------CIFAR10 RESNET 20 RESULTS

From the figure, the following observations can be made:

%\paragraph{}\label{exp1:obv1} \vspace{-9pt}
\textbf{Observation 1: Transferability of loss gradient based attacks is poor.} As can be seen, transferability of FGSM, CW, and UAP is poor. CW attack, which has 100\% efficiency when applied on the same network, transfers very poorly. One of the reasons for the observed poor transferability of these attacks is that the gradient of the loss function (with respect to the input) of the networks at different bitwidths do not align with each other \cite{impact_low_bit}. Loss gradient based attacks create adversarial examples by adding perturbations in the direction of the loss gradient. Thus, misalignment of gradient between source and target means that the perturbation direction transferred from source may not be able to cause the same effect at the target. The lower the cosine similarity between the loss gradients, the lower is the transferability \cite{understand_enhance, demontis_why_adv_transfer, impact_low_bit}. By comparison, UAP has better transferability which is noticed in observation 3.

Another reason for the poor transferability could be due to the \emph{quantization shift} phenomenon \cite{impact_low_bit, to_compress}. When networks are quantized, weight and activation values are grouped into the same bucket, this impacts how the attack performs when bitwidth is changed. For instance, when adversarial examples are crafted in a quantized network, an algorithm may assume that weight values associated with certain nodes are similar, but when moving the attack to a full-precision network, one weight value may be more than the other. This might end up activating unintended nodes, thus making the sample unsuccessful. This also explains why a powerful attack like CW fails to transfer. Since the attack relies on creating differential activation between the logits, when the attack is transferred to another network, the attack might not be able to achieve the same activation difference between the true class and adversarial class. Hence, the change in bitwidth between source and target hampers the effectiveness of the attack.

%Likewise, when creating examples in a FP network, an algorithm may leverage weights which are higher than others in order to overdrive certain activations which eventually leads to a misclassification. When this example is transferred to a quantized network, the adversarial effect may be lost as the quantization of weights can cause multiple weight values to be quantized to the same level as the important weights that were used by the attack when crafting the example; similarly, activation quantization can cause multiple activations to be quantized to the same value.  

We also noticed that the transferability of CW attack can be improved by increasing the value of $\kappa$. This was also observed in the original paper \cite{cw_attack} as well as by other studies \cite{impact_low_bit}. However, in our experiments, when the value was increased beyond $\kappa=5$, most of the images became unrecognizable. Since one of the requirements for a sample to be adversarial is that it should be recognizable to humans, we did not increase $\kappa$ beyond this value. 

%\paragraph{}\label{exp1:obv2} \vspace{-9pt}
\textbf{Observation 2: JSMA has poor transferability.} This can again be explained in terms of quantization shift. JSMA crafts adversarial examples by manipulating individual features in an image based on how effective that feature is in producing misclassification. The effectiveness of each feature is based on individual parameter and activation values of that network. Due to quantization shift, the same features may no longer be sensitive on a different network with different bitwidth. This explains why JSMA has better performance when the attack is applied on the same network as it is able to find features that are sensitive enough to cause misclassification for that network. However, when the created sample is transferred to another network with different bitwidth, the same perturbations are no longer able to produce similar change in activations due to the target network being in full-precision or having different discrete levels of activation and weight values.

%\label{exp1:obv3} \vspace{-9pt}
\textbf{Observation 3: The Boundary Attack has poor transferability.} As can be seen, the transferability of the Boundary Attack is poor. The target networks are highly robust against the attack even though the same examples work with almost 100\% success rate at the source. This can be explained with how the Boundary Attack crafts adversarial examples and the resulting direction of the perturbations.

Adversarial examples exist in a broad contiguous regions in input space; thus, rather than an exact example, the direction of perturbation is important for transferability \cite{exp_harness}. The perturbation directions that lead to the adversarial sub-space are called the adversarial directions and the examples created on source are transferable if their adversarial directions align with adversarial directions of the black-box target model \cite{space_for_adv}. Thus, aligning the perturbation with the adversarial direction is a reliable way of not only fooling the source but also the target model. These adversarial directions point away from the original image as the perturbations are normally added to the original image. However, in case of the Boundary Attack, perturbations are added to a random image, moving it towards the original image. The perturbed image thus has perturbation direction towards the original class boundary which is opposite to the adversarial direction. This explains the poor transferability of the attack. 

%\paragraph{}\label{exp1:obv4} \vspace{-9pt}
\textbf{Observation 4: UAP has better transferability than FGSM when the allowed perturbation is high.} Although UAP algorithm internally uses FGSM to compute perturbations for individual images (Section \ref{sec:craft_adv}), it performs better than FGSM itself when the allowed magnitude of perturbation $\xi$ is high.

\begin{figure} [!ht]
\centering
\begin{subfigure}{.24\textwidth}
  \centering
  \includegraphics[width=.95\linewidth]{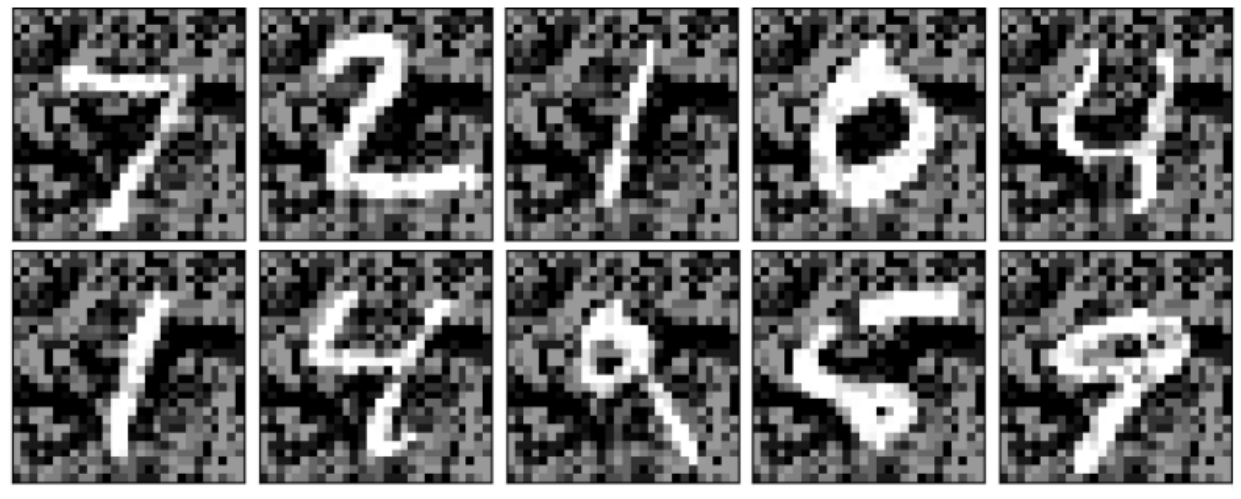}
  \caption{}
  \label{fig:mnist_eps_06}
\end{subfigure}%
\begin{subfigure}{.24\textwidth}
  \centering
  \includegraphics[width=.95\linewidth]{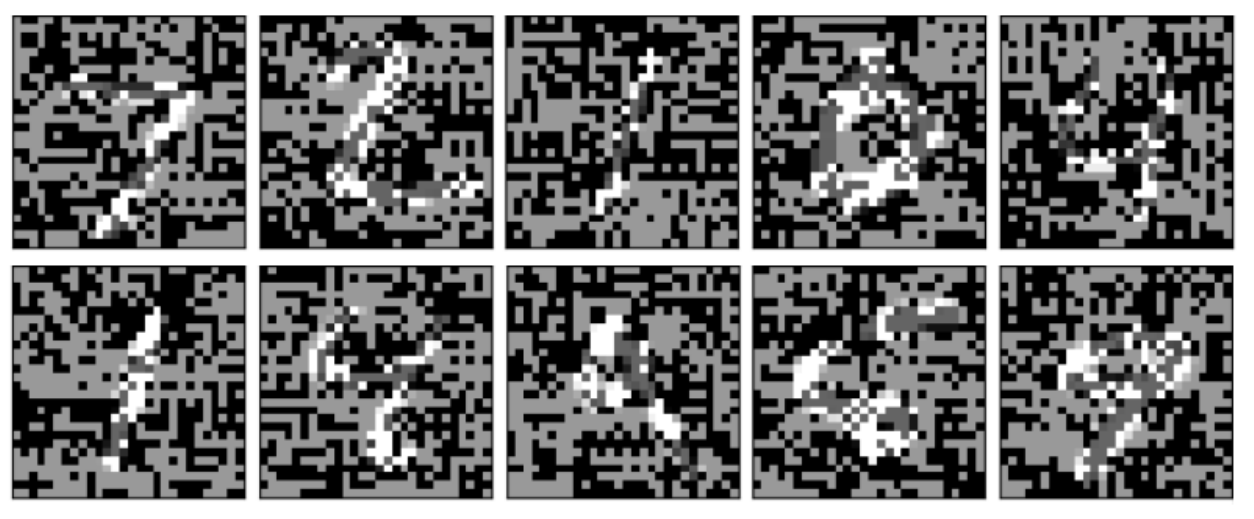}
  \caption{}
  \label{fig:mnist_uap_xi_06}
\end{subfigure}
\captionsetup{singlelinecheck=off}
\caption[Adversarial examples generated on FP Mnist A model using UAP and FGSM for the same magnitude of allowed distortion.]{Adversarial examples generated on FP Mnist A model using: \begin{inparaenum} [(a)] \item UAP with $\xi = 0.6$.  \item FGSM with  $\varepsilon = 0.6$. Both images are first 10 images from the MNIST dataset. \end{inparaenum}}
\Description{Adversarial examples generated on FP Mnist A model using: (a) UAP with $\xi = 0.6$. (b) FGSM with  $\varepsilon = 0.6$. Both images are first 10 images from the MNIST dataset.}
\label{fig:uapvsfgsmcompare}
\end{figure}

UAP allows to increase the maximum allowed distortion to higher values than the internal algorithm while still keeping the adversarial image recognizable. Figure~\hyperref[tbl:Models_cifar_high_cap_arch]{2} compares adversarial images generated using UAP and FGSM on the FP Mnist A model by taking the same value of allowed distortion ($\varepsilon$ for FGSM and $\xi$ for UAP). As can be seen, the images still remain recognizable in the case of UAP, whereas the images are completely distorted for FGSM.

\noindent
%\paragraph{}\label{summary1} \vspace{-9pt}
\textbf{Summary}  

\begin{itemize}

% \item \label{exp1:summary_1} \emph{Attack performance on source is not an indication of transferability.} The Boundary Attack has very high success rate on the source network but the transferability is poor. Other attacks like FGSM which have less success on the source network depict comparatively better transferability.

\item \label{exp1:summary_2} \emph{Quantization reduces the transferability of attacks.} Quantization shift and misalignment of loss gradients between variable bitwidth networks result in poor transferability. Similar observations were made in \cite{impact_low_bit, to_compress}. The experiments in this section extend this finding to show that the quantization shift also affects the transferability of algorithms like JSMA that leverage individual features to produce misclassification. Further, the experiments on the Boundary Attack show that search based algorithms might not produce transferable examples due to the direction of perturbation. 

\item \label{exp1:summary_3} \emph{At higher distortions, UAPs are more transferable than the attack used to craft it.} It was found that the UAP allows to add more distortion than the algorithm used internally to craft it. Thus, the performance of UAP may be enhanced to a greater extent than that of the internal algorithm.
\end{itemize}

\subsection{Transferability of adversarial examples when considering model-related properties} \label{sub:exp2}
To explore how model-related properties affect transferability among networks of different bitwidths, adversarial examples created at source were transferred to target networks which not only had different bitwidths but also different \emph{model capacity} and \emph{model architecture} than the source model. The model capacity here is quantified in terms of number of parameters in the model. 

% To observe how model capacity affects transferability, additional CIFAR10 models each of different capacity were trained. More specifically, Resnet32, Resnet44, and their quantized versions were trained. Details on these models are presented in the Appendix. Using FP Resnet44 and its quantized versions as targets, attacks were transferred from FP and quantized versions of Resnet20 and Resnet32. Similarly, to study how model architecture affects transferability, a 7-layer CNN, named as Cifar A for easier reference, was trained on the CIFAR10 dataset. Adversarial examples crafted on FP and quantized versions of Resnet20 was then transferred to FP and quantized versions of Cifar A. The transferability results from both model-capacity variation and architecture variation are quite similar, thus to save space only the results from different architecture transfers are presented in Figure \ref{fig:transferability_res_resnet20_diffarch}. The results from different model-capacity transfers are presented in the Appendix.%

\begin{table}[htpb]
\caption[Test set accuracy of the full-precision (FP) and quantized versions of high-capacity variants of Mnist A model.]{Test set accuracy of the full-precision (FP) and quantized versions of high-capacity variants of Mnist A model.  Mnist A has 414K parameters, while Mnist B and Mnist C have 836K and 1.7M parameters, respectively.}
\Description{Test set accuracy of the full-precision (FP) and quantized versions of high-capacity variants of Mnist A model.  Mnist A has 414K parameters, while Mnist B and Mnist C have 836K and 1.7M parameters, respectively.}
\label{tbl:Models_mnist_high_cap}
\centering
	\begin{tabular}	{m{0.072\textwidth}  m{0.034\textwidth}  m{0.034\textwidth} m{0.034\textwidth}  m{0.034\textwidth} m{0.034\textwidth}  m{0.043\textwidth}  m{0.043\textwidth}}	
		\toprule
		\textbf{Model ID}& \textbf{FP} & \textbf{1 bit} & \textbf{2 bit} & \textbf{4 bit} & \textbf{8 bit} & \textbf{12 bit} & \textbf{16 bit}\\ 
		\toprule	
		%-------------------------------
		Mnist B  & 0.994  & 0.993  & 0.993 & 0.991  & 0.994 & 0.992  & 0.994\\ 	
		\midrule
		Mnist C & 0.993  & 0.994  & 0.990 & 0.991  & 0.993 & 0.992  & 0.993\\ 
		\bottomrule
	\end{tabular}	
\end{table}

\begin{table}[htpb]
\caption[Test set accuracy of the full-precision (FP) and quantized versions of various ResNets and CNN models used.]{Test set accuracy of the full-precision (FP) and quantized versions of various ResNets and CNN models used. Resnet20 has 269K parameters, while Resnet32, Resnet44, and Cifar A have 464K, 658K, and 4.5M parameters, respectively.}
\Description{Test set accuracy of the full-precision (FP) and quantized versions of various ResNets and CNN models used. Resnet20 has 269K parameters, while Resnet32, Resnet44, and Cifar A have 464K, 658K, and 4.5M parameters, respectively.}
\label{tbl:Models_cifar_high_cap_arch}
\centering
	\begin{tabular}	{m{0.072\textwidth}  m{0.034\textwidth}  m{0.034\textwidth} m{0.034\textwidth}  m{0.034\textwidth} m{0.034\textwidth}  m{0.043\textwidth}  m{0.043\textwidth}}		
		\toprule
		\textbf{Model ID}& \textbf{FP} & \textbf{1 bit} & \textbf{2 bit} & \textbf{4 bit} & \textbf{8 bit} & \textbf{12 bit} & \textbf{16 bit}\\ 
		\toprule	
		%-------------------------------
		Resnet32  & 0.901  & 0.862  & 0.866 & 0.864  & 0.861 & 0.845  & 0.850\\ 	
		\midrule
		Resnet44 & 0.911  & 0.868  & 0.873 & 0.877  & 0.860 & 0.854  & 0.861\\ 
		\midrule
		Cifar A & 0.894  & 0.846  & 0.864 & 0.866  & 0.862 & 0.866  & 0.868\\
		\bottomrule
	\end{tabular}	
\end{table}

%-------------------------------------------------CIFAR10 RESNET_20 TO RESNET_44 RESULTS
\begin{figure*}[!htp]
%hp = put figure "here" or on "top" or in a page for "only float"
\centering
%---FGSM_005 RESNET 20------
\begin{subfigure}{.34\textwidth}
  \centering
  \includegraphics[width=0.99\linewidth]{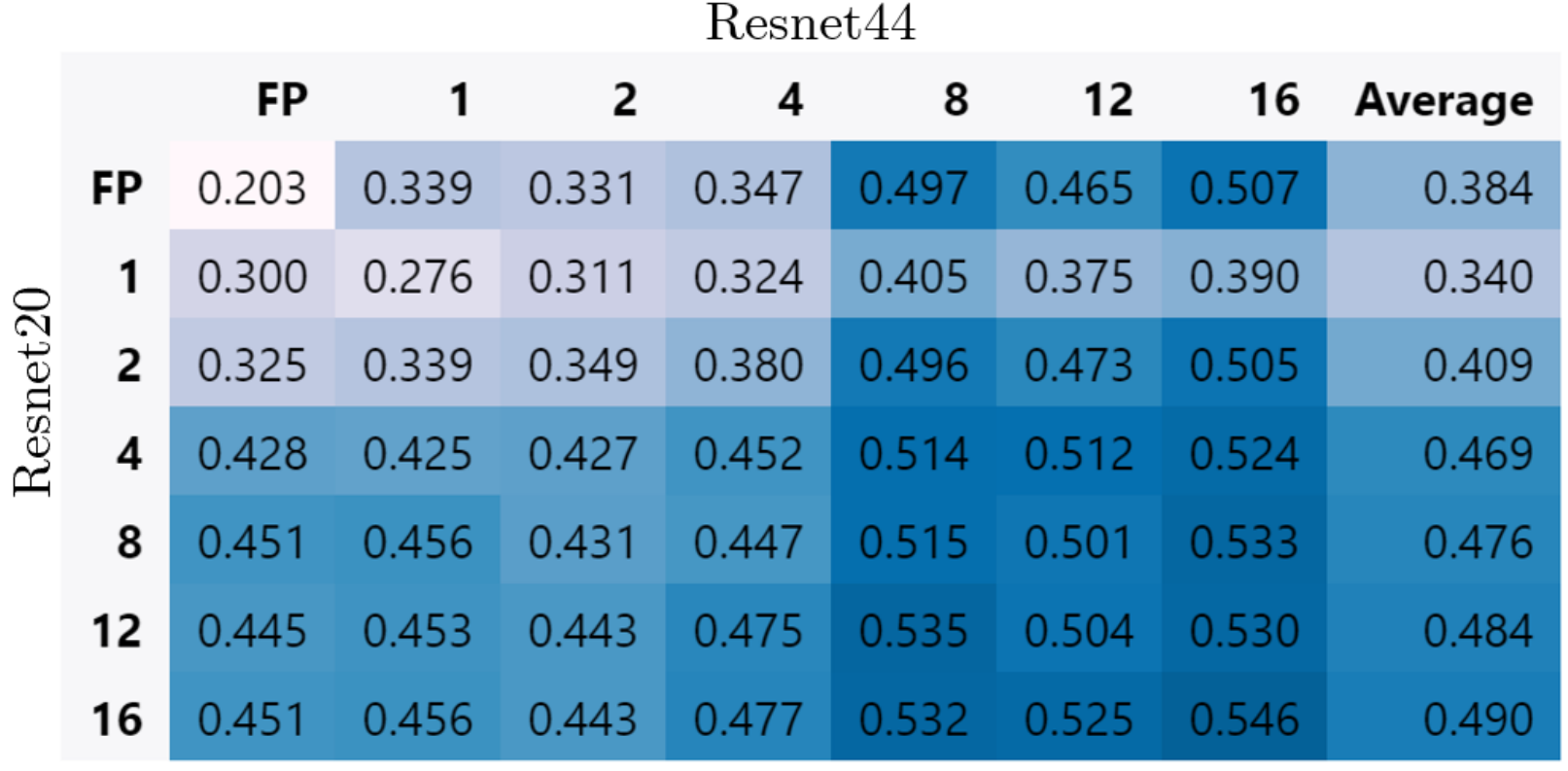}
  \caption*{FGSM ($\varepsilon = 0.05$)}
  \label{fig:cifar_FGSM_2}  
\end{subfigure}\hspace{5mm}%
%---FGSM_005 RESNET 32------
\begin{subfigure}{.34\textwidth}
  \centering
  \includegraphics[width=0.99\linewidth]{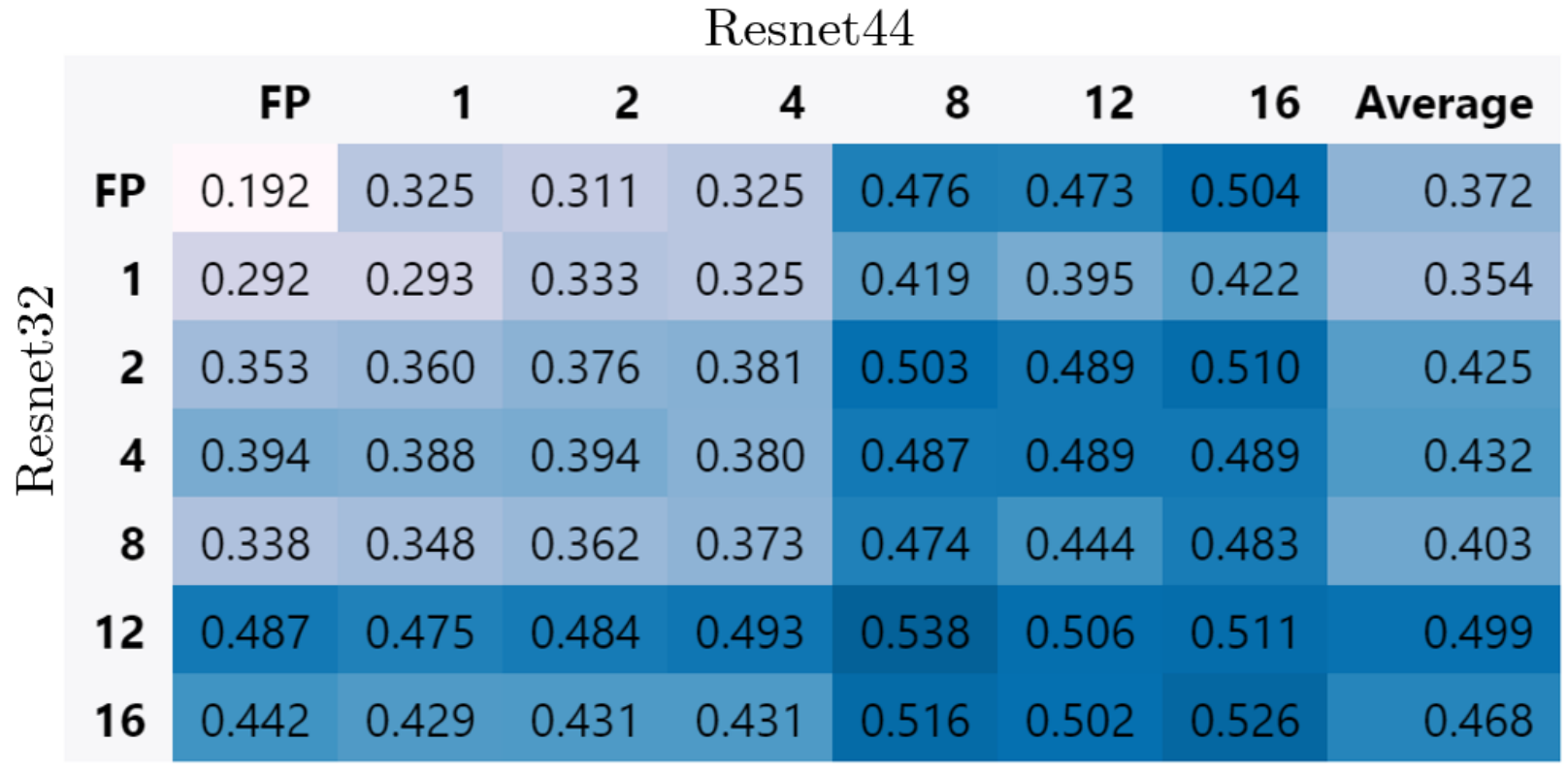}
  \caption*{FGSM ($\varepsilon = 0.05$)}
  \label{fig:cifar_FGSM_2}  
\end{subfigure}

%---JSMA_T_03_G_005 RESNET 20------
\bigskip
\begin{subfigure}{.34\textwidth}
  \centering
  \includegraphics[width=0.99\linewidth]{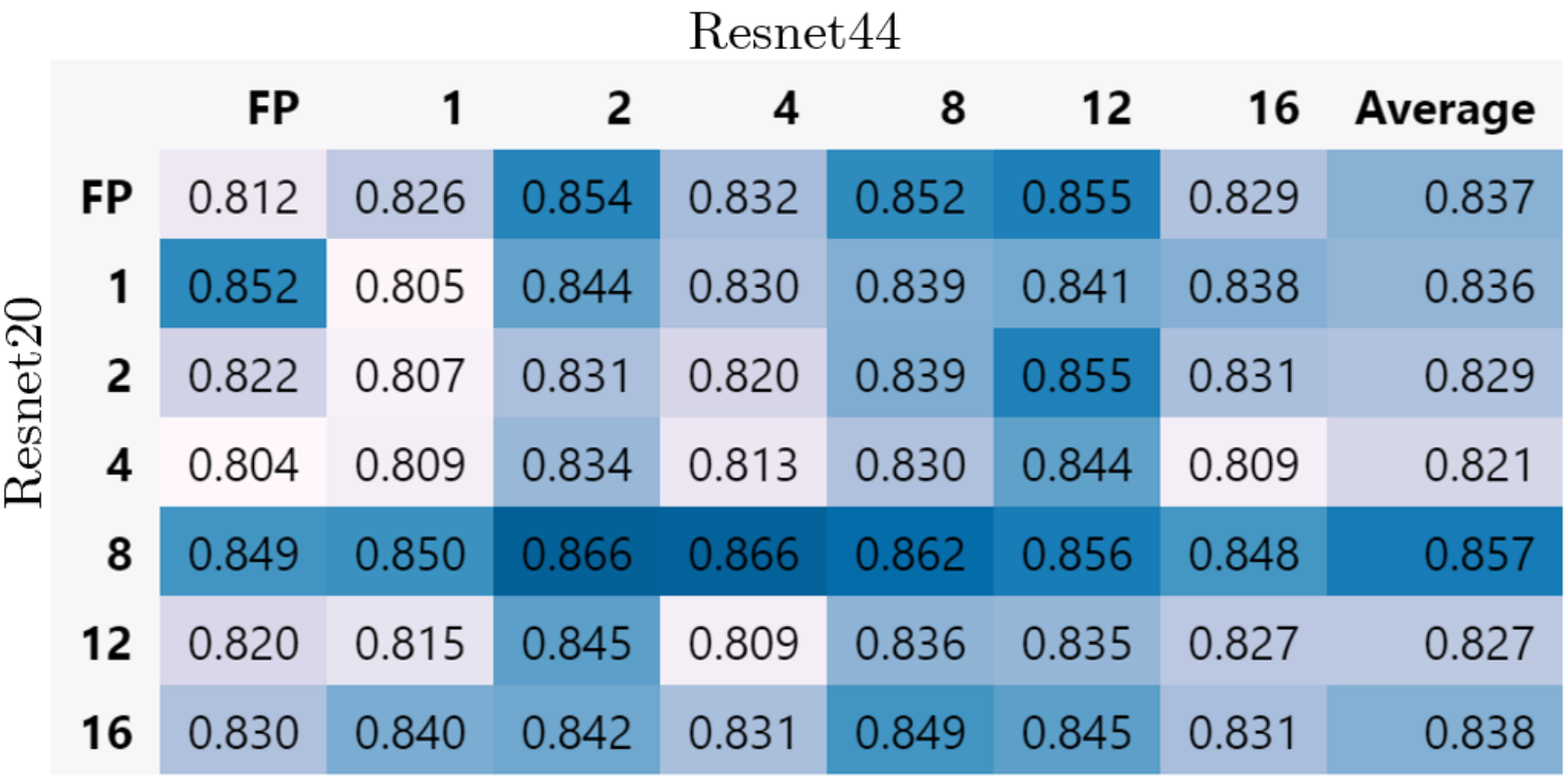}
  \caption*{JSMA ($\theta = 0.3$, $\gamma = 5\%$)}
  \label{fig:cifar_JSMA_2}
\end{subfigure}\hspace{5mm}%
%---JSMA_T_03_G_005 RESNET 32------
\begin{subfigure}{.34\textwidth}
  \centering
  \includegraphics[width=0.99\linewidth]{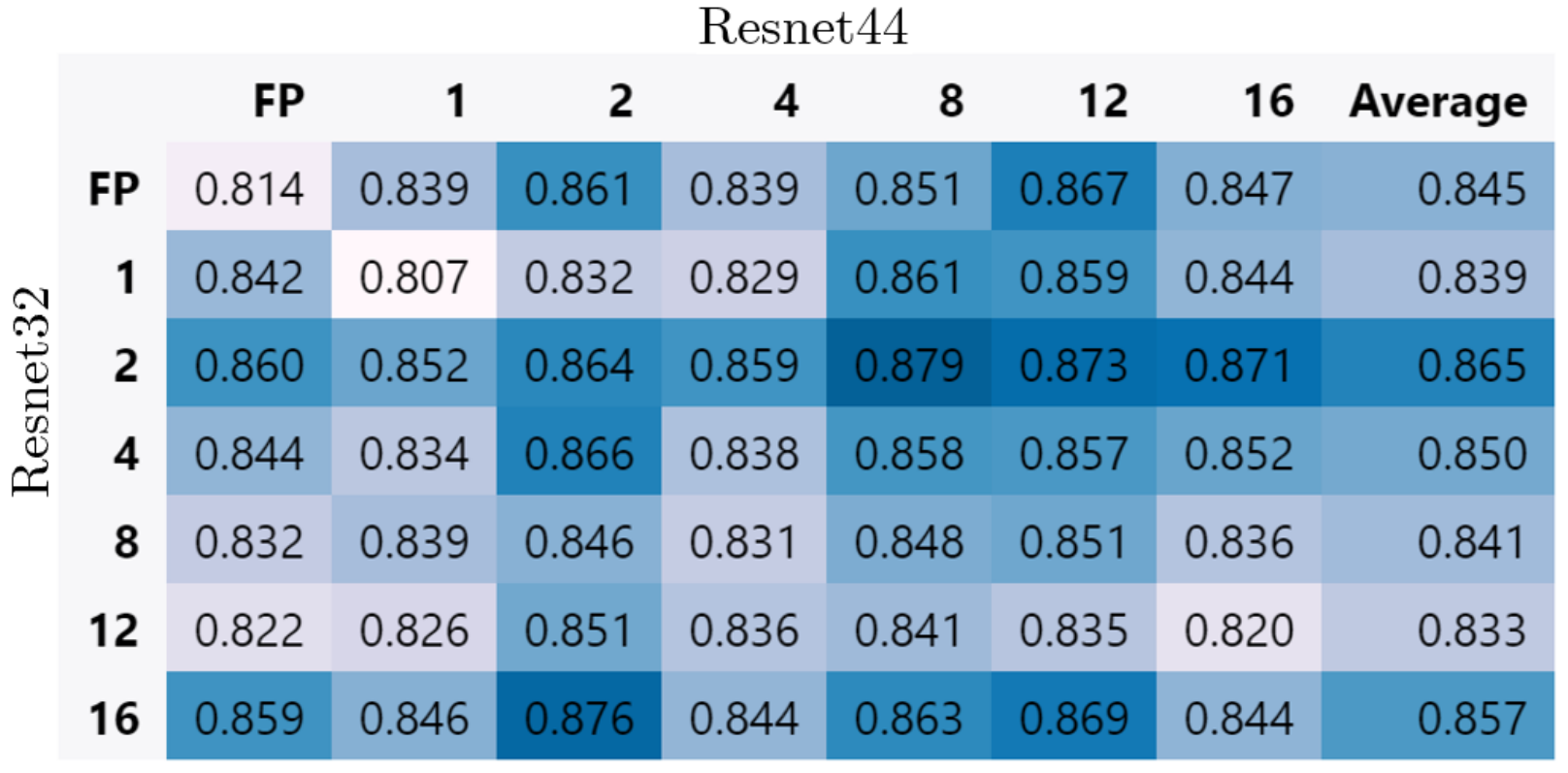}
  \caption*{JSMA ($\theta = 0.3$, $\gamma = 5\%$)}
  \label{fig:cifar_JSMA_2}
\end{subfigure}

%---UAP_EP_001_XI_01 RESNET 20------
\bigskip
\begin{subfigure}{.34\textwidth}
  \centering
  \includegraphics[width=0.99\linewidth]{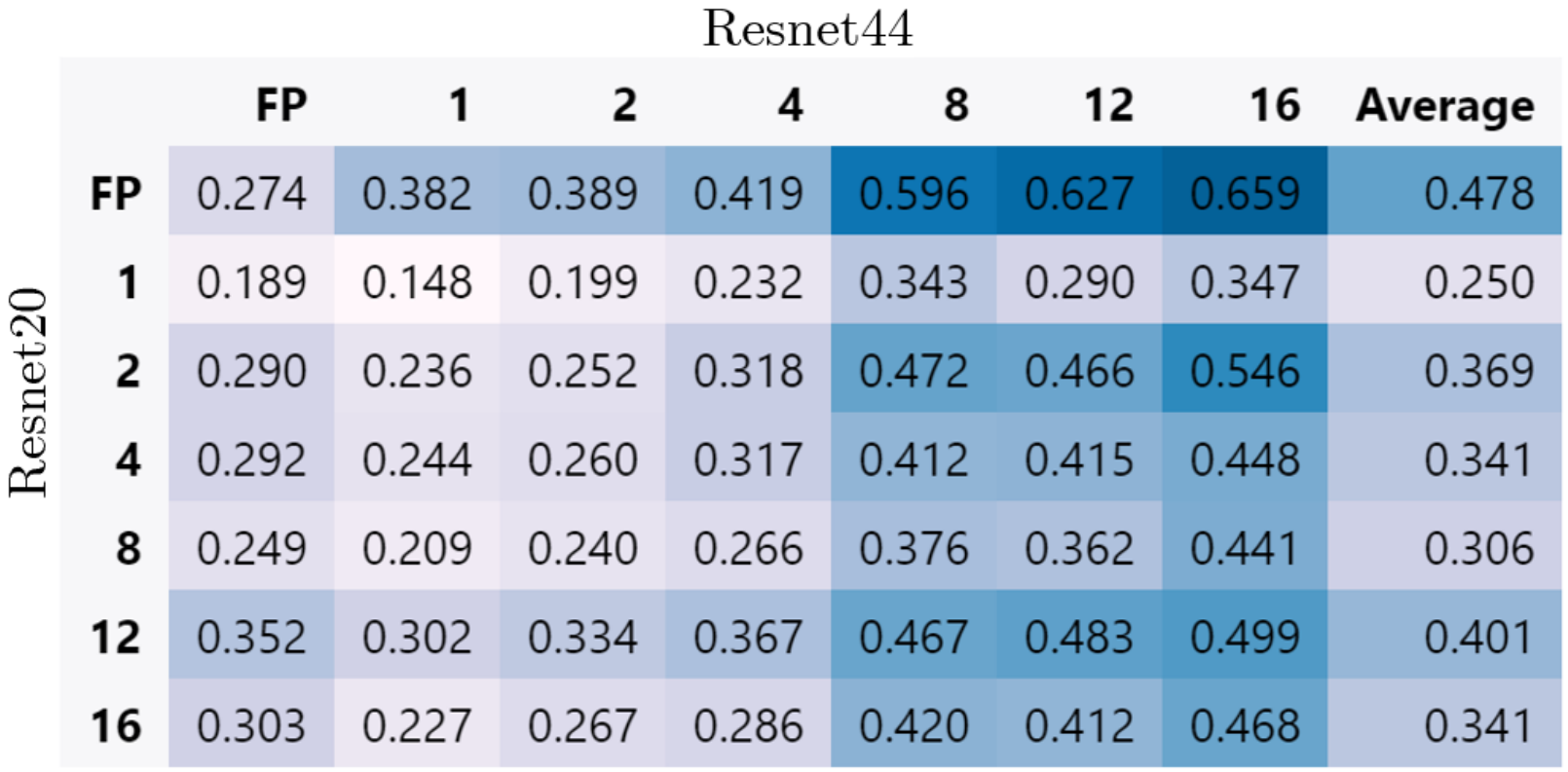}
  \caption*{UAP ($\varepsilon = 0.01$, $\xi = 0.1$)}
  \label{fig:cifar_UAP_2}
\end{subfigure}\hspace{5mm}%
%---UAP_EP_001_XI_01 RESNET 32------
\begin{subfigure}{.34\textwidth}
  \centering
  \includegraphics[width=0.99\linewidth]{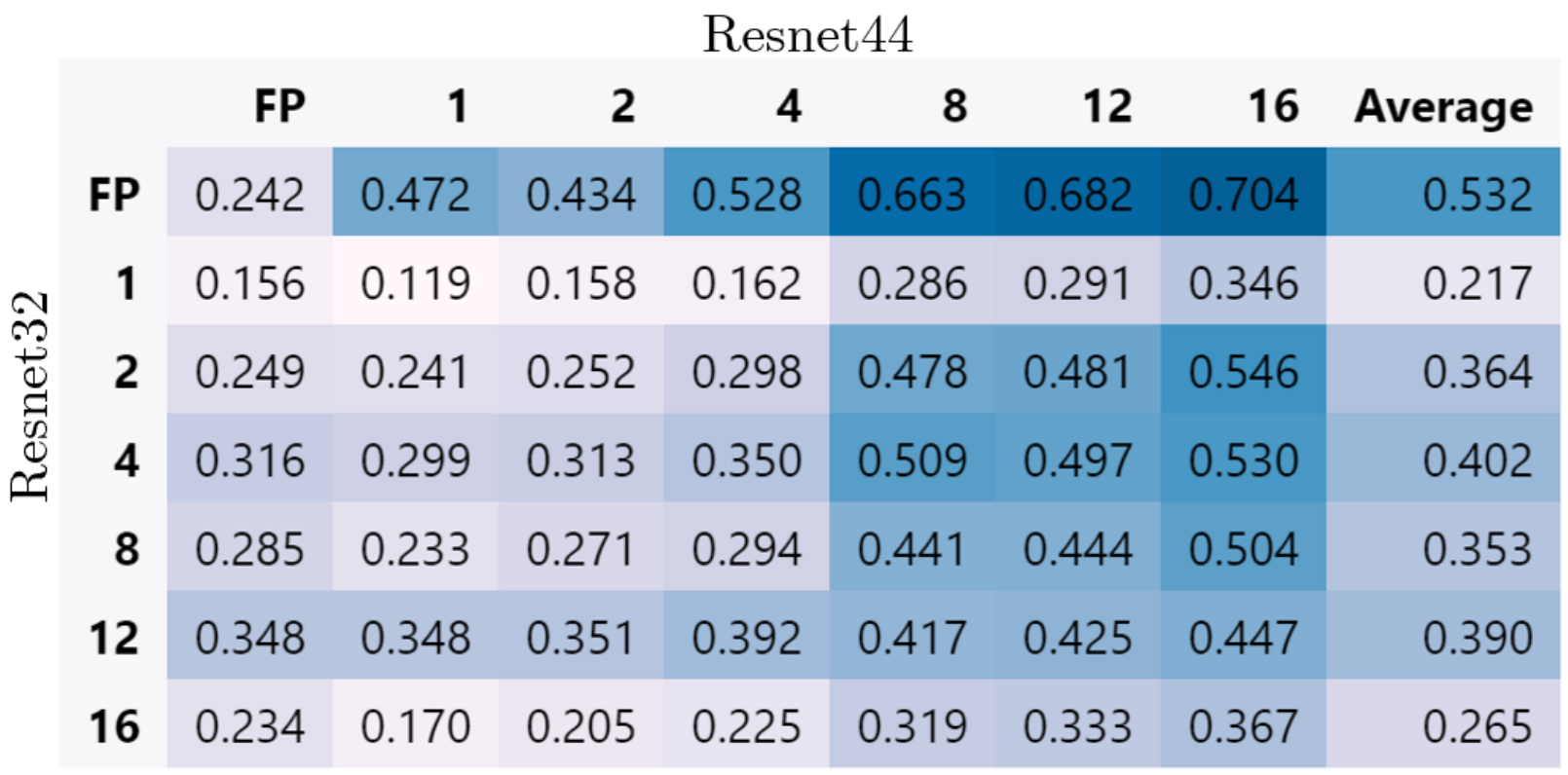}
  \caption*{UAP ($\varepsilon = 0.01$, $\xi = 0.1$)}
  \label{fig:cifar_UAP_2}
\end{subfigure}

%---CW_K_5 RESNET 20------
\bigskip
\begin{subfigure}{.34\textwidth}
  \centering
  \includegraphics[width=0.99\linewidth]{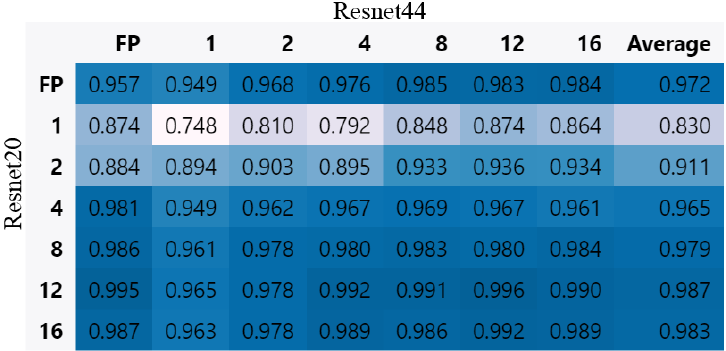}
  \caption*{CW ($\kappa = 5$, $i = 25$, $b_{s} = 20$)}
  \label{fig:cifar_CW_1}   
\end{subfigure}\hspace{5mm}%
%---CW_K_5_ITT RESNET 32------
\begin{subfigure}{.34\textwidth}
  \centering
  \includegraphics[width=0.99\linewidth]{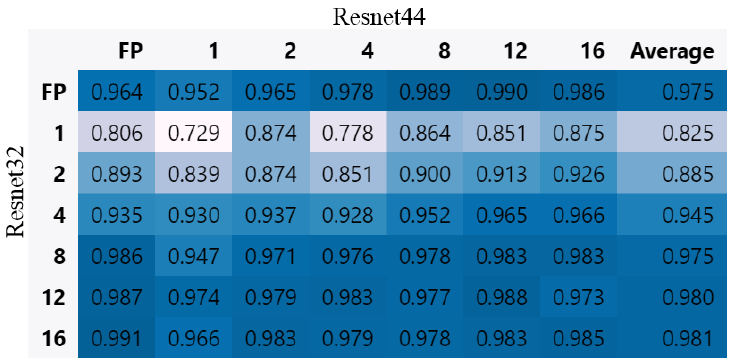}
  \caption*{CW ($\kappa = 5$, $i = 25$, $b_{s} = 20$)}
  \label{fig:cifar_CW_1}  
\end{subfigure}

%---BA_12_ITT RESNET 20------
\bigskip
\begin{subfigure}{.34\textwidth}
  \centering
  \includegraphics[width=0.99\linewidth]{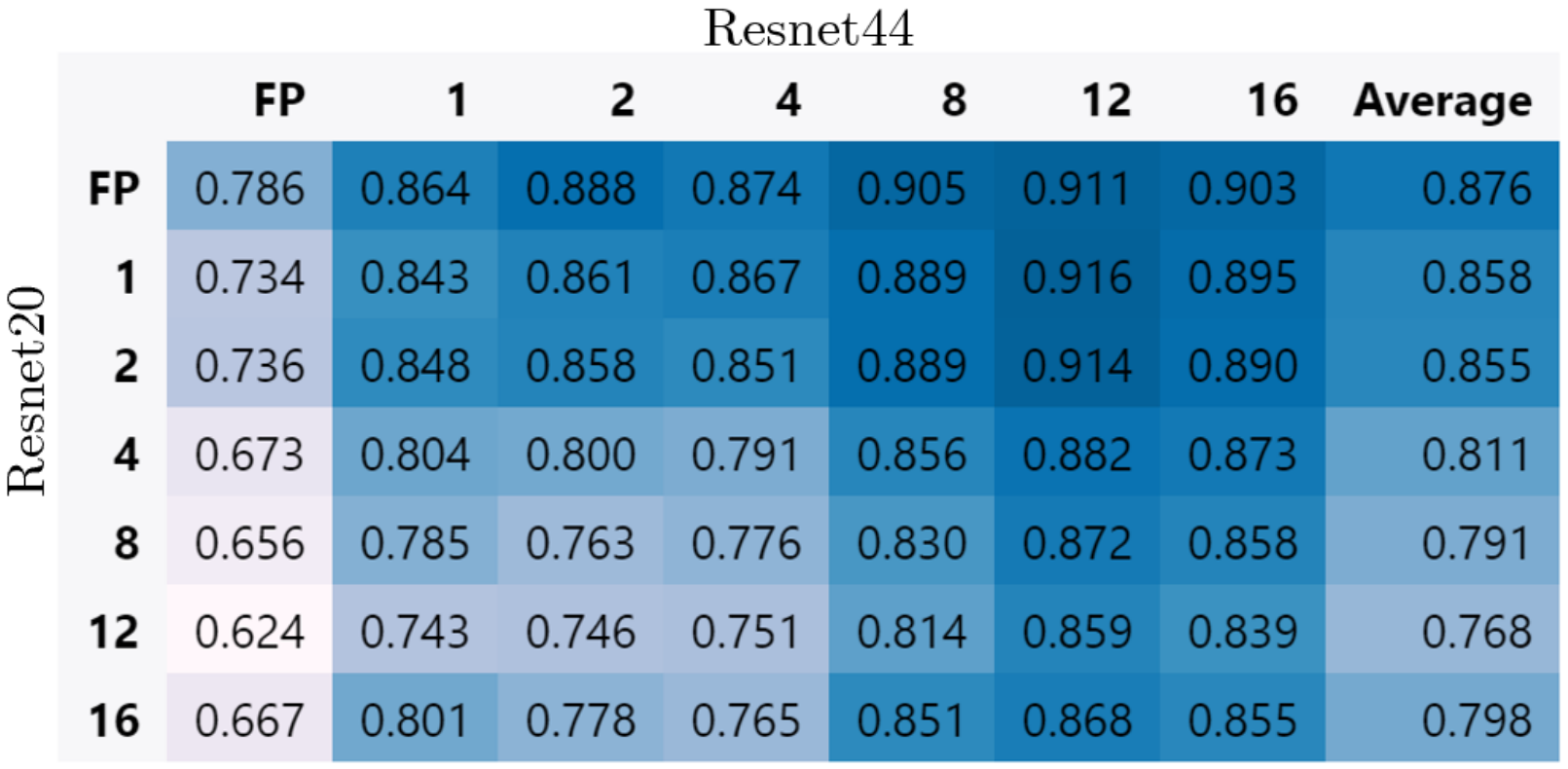}
  \caption*{BA ($i = 12$)}
  \label{fig:cifar_BA_1}  
\end{subfigure}\hspace{5mm}%
%---BA_12_ITT RESNET 32------
\begin{subfigure}{.34\textwidth}
  \centering
  \includegraphics[width=0.99\linewidth]{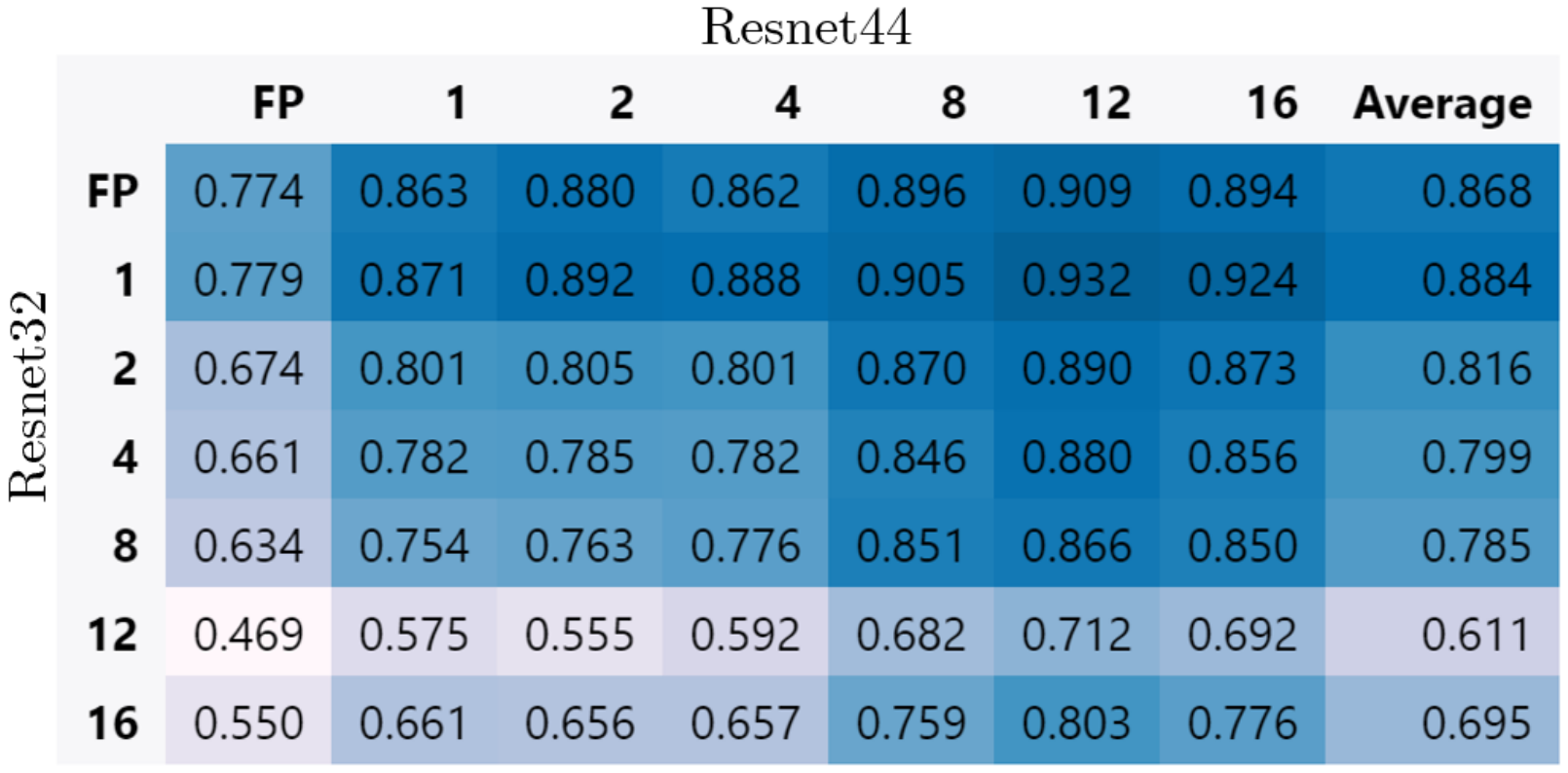}
  \caption*{BA ($i = 12$)}
  \label{fig:cifar_BA_1}  
\end{subfigure}

\captionsetup{singlelinecheck=off}
\caption[Transferability of adversarial attacks among CIFAR10 models when the source and target networks differ in capacity.]{Transferability of adversarial attacks when the source and target networks differ in capacity. The five matrices on the left column depict the transferability of each of the five attacks when the source networks are different bitwidth versions of Resnet20. The matrices on the right column depict the transferability of the same attacks when the source networks are different bitwidth versions of Resnet32. The target networks are FP Resnet44 and its quantized versions in all cases.}
\Description{Transferability of adversarial attacks when the source and target networks differ in capacity. The five matrices on the left column depict the transferability of each of the five attacks when the source networks are different bitwidth versions of Resnet20. The matrices on the right column depict the transferability of the same attacks when the source networks are different bitwidth versions of Resnet32. The target networks are FP Resnet44 and its quantized versions in all cases.}
\label{fig:transferability_res_resnet20_diffcap}
\end{figure*}

\begin{figure*}[!htp]
%hp = put figure "here" or on "top" or in a page for "only float"
\centering
%---FGSM_005 RESNET 20------
\begin{subfigure}{.34\textwidth}
  \centering
  \includegraphics[width=0.99\linewidth]{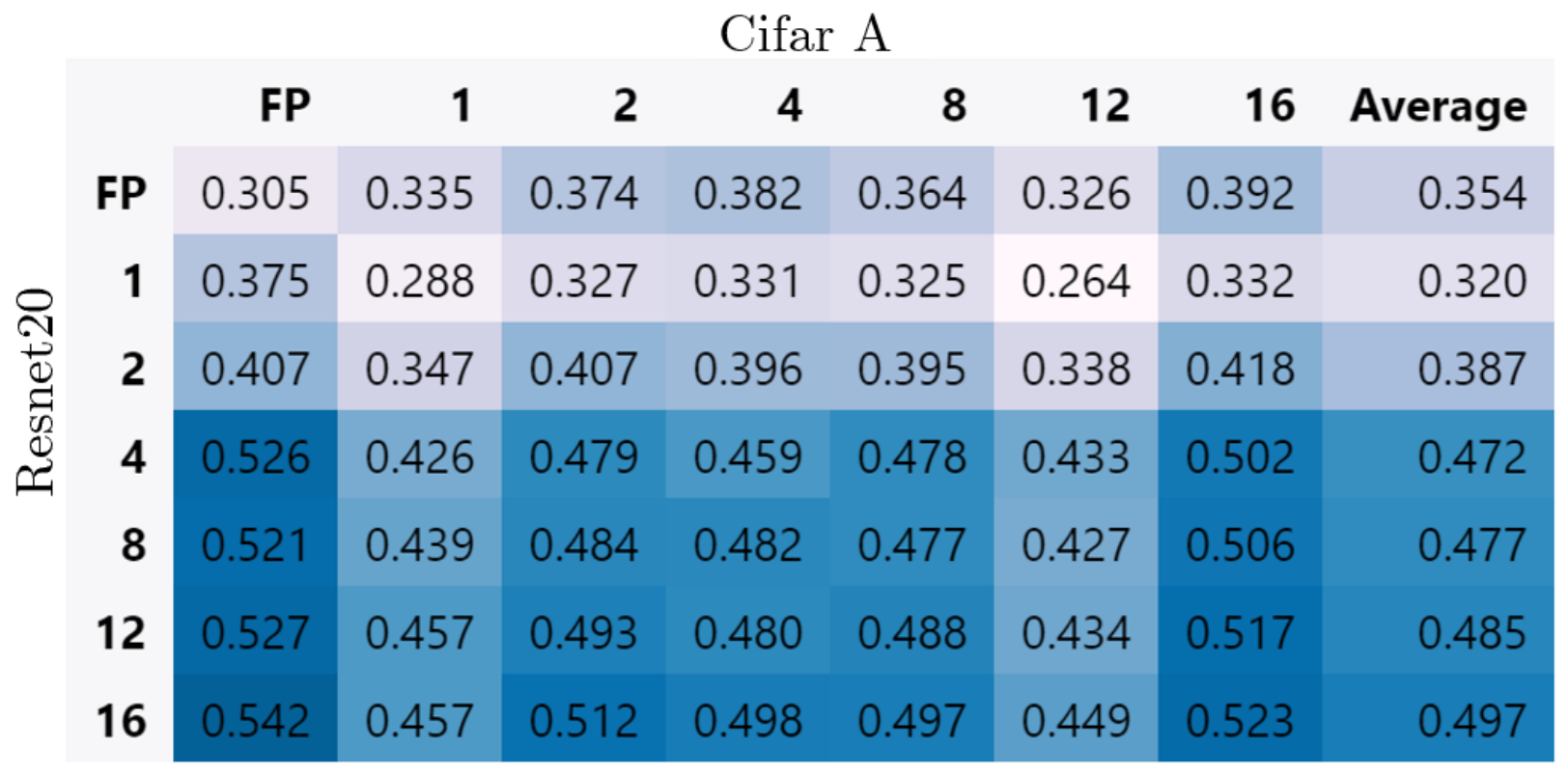}
  \caption*{FGSM ($\varepsilon = 0.05$)}
  \label{fig:cifar_FGSM_arch}  
\end{subfigure}%
%---JSMA_T_03_G_005 RESNET 20------
\begin{subfigure}{.34\textwidth}
  \centering
  \includegraphics[width=0.99\linewidth]{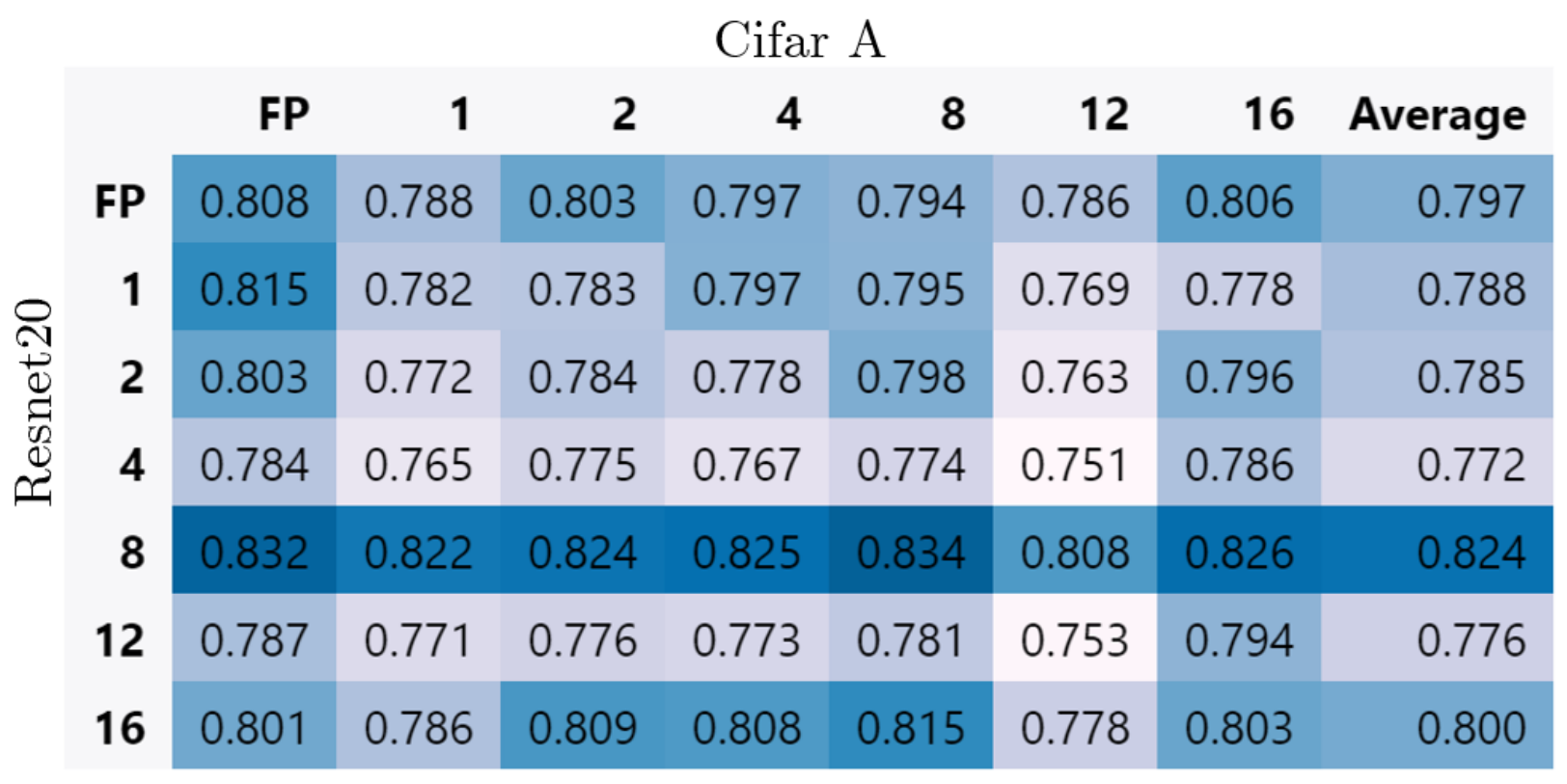}
  \caption*{JSMA ($\theta = 0.3$, $\gamma = 5\%$)}
  \label{fig:cifar_JSMA_arch}
\end{subfigure}%
%---UAP_EP_001_XI_01 RESNET 20------
\bigskip
\begin{subfigure}{.34\textwidth}
  \centering
  \includegraphics[width=0.99\linewidth]{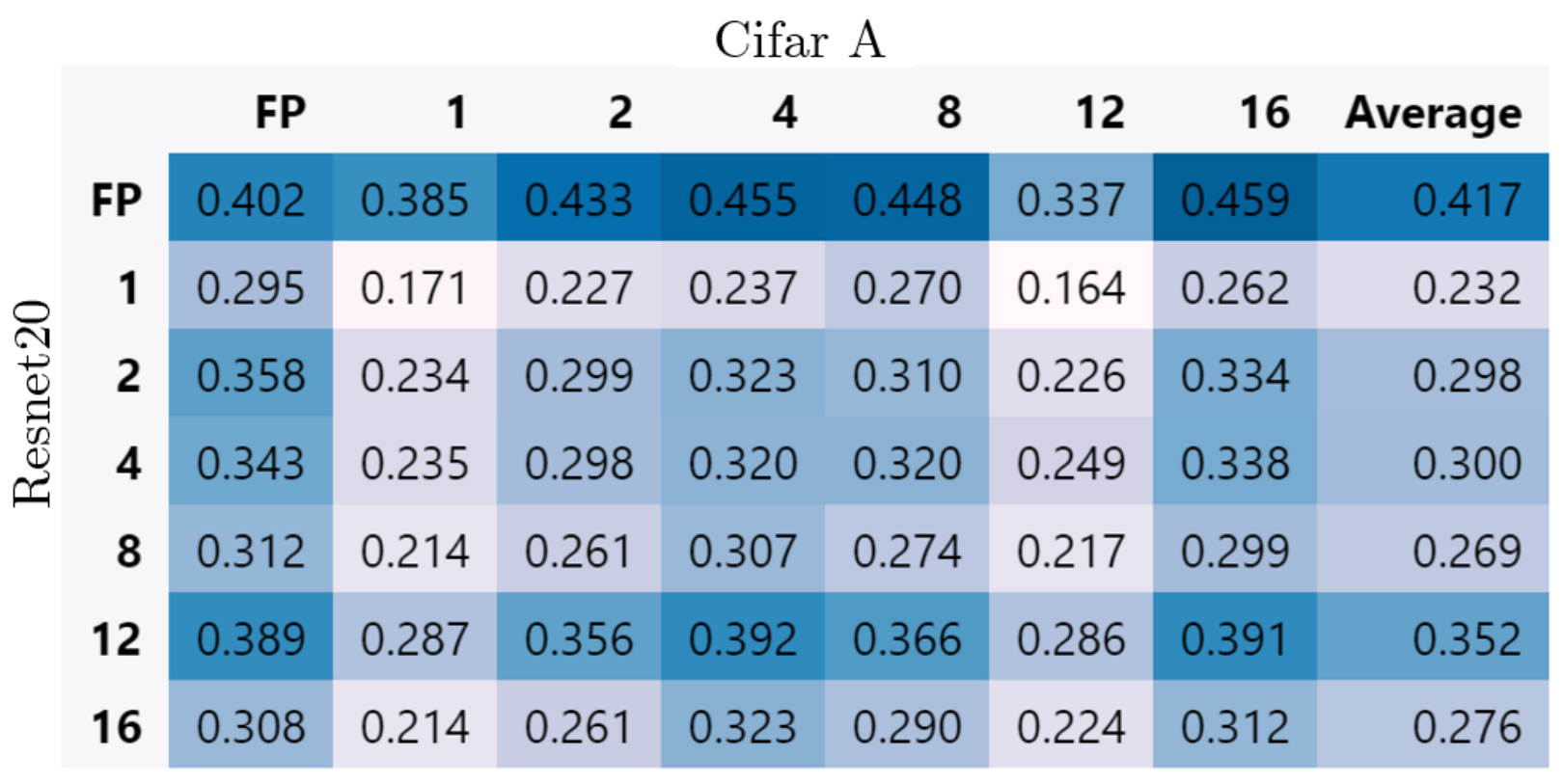}
  \caption*{UAP ($\varepsilon = 0.01$, $\xi = 0.1$)}
  \label{fig:cifar_UAP_arch}
\end{subfigure}
%---CW_BSS_20_K_5_ITT_30------
\begin{subfigure}{.34\textwidth}
  \centering
  \includegraphics[width=0.99\linewidth]{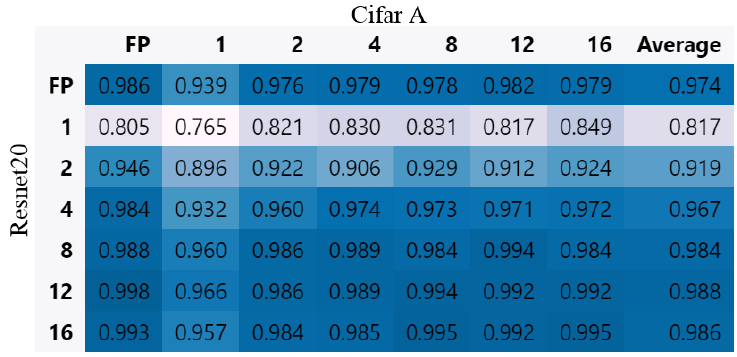}
  \caption*{CW ($\kappa = 5$, $i = 25$, $b_{s} = 20$)}
  \label{fig:cifar_CW_arch}
\end{subfigure}\hspace{2.5mm}%
%---BA_12_ITT RESNET 20------
\begin{subfigure}{.34\textwidth}
  \centering
  \includegraphics[width=0.99\linewidth]{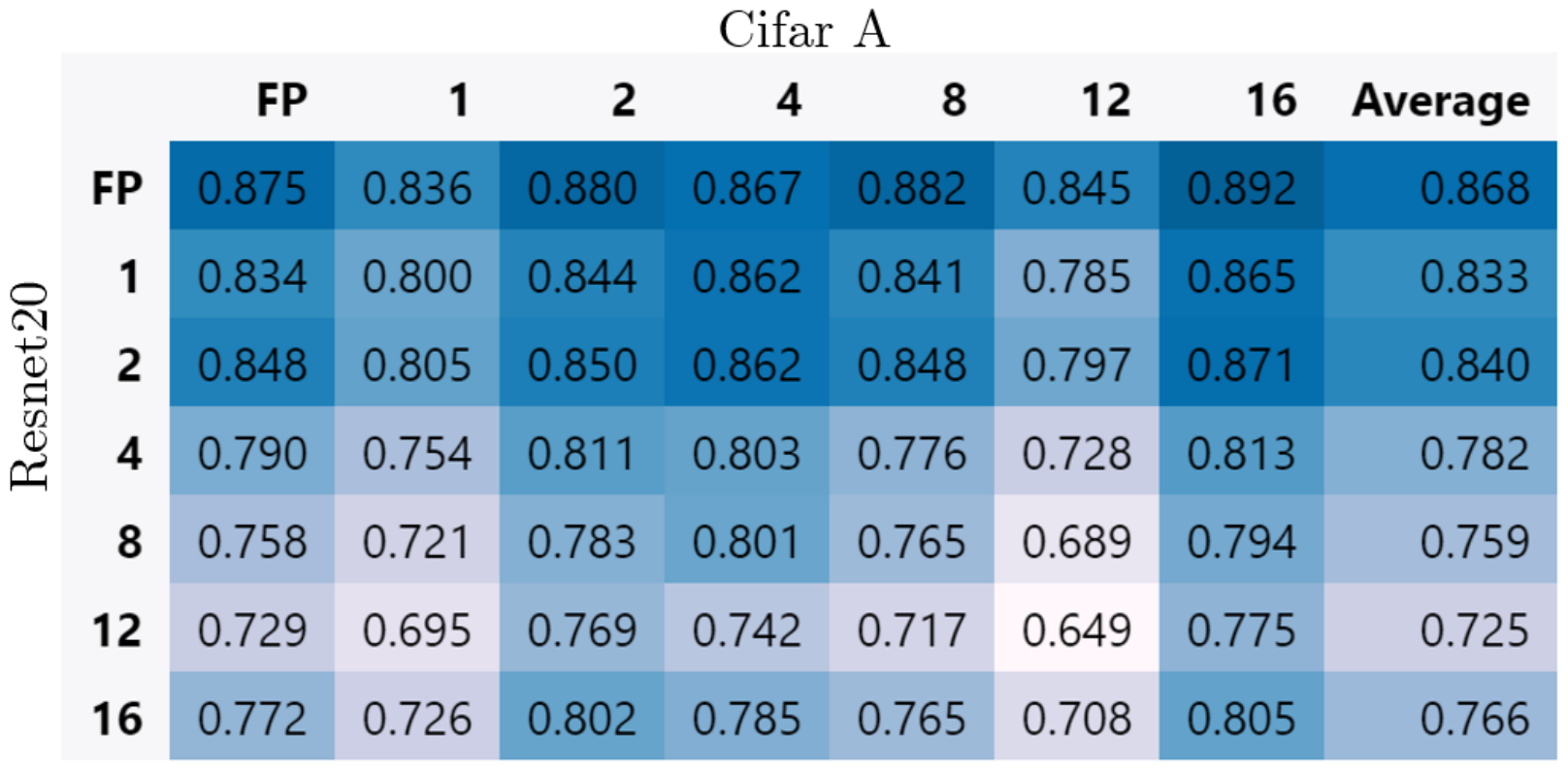}
  \caption*{BA ($i = 12$)}
  \label{fig:cifar_BA_arch}  
\end{subfigure}%
\captionsetup{singlelinecheck=off}
\caption[Transferability of adversarial attacks among CIFAR10 models when the source and target networks differ in architecture.]{Transferability of adversarial attacks when the source and target networks differ in architecture. The source networks are different bitwidth versions of Resnet20 and the target networks are different bitwidth versions of Cifar A.}
\Description{Transferability of adversarial attacks when the source and target networks differ in architecture. The source networks are different bitwidth versions of Resnet20 and the target networks are different bitwidth versions of Cifar A.}
\label{fig:transferability_res_resnet20_diffarch}
\end{figure*}

% Based on the data in Figure \ref{fig:transferability_res_resnet20_diffarch}, the following observations can be made:

Additional CIFAR10 and MNIST models of different capacities were trained. More specifically, Resnet32, Resnet44, and their quantized versions were trained for CIFAR10 and by increasing the number of channels for all convolution layers of Mnist A by 100\% and 300\%, Mnist B and Mnist C, respectively, were trained for MNIST. Tables~\ref{tbl:Models_mnist_high_cap} and \ref{tbl:Models_cifar_high_cap_arch} show test accuracies for quantized and FP versions of both MNIST and CIFAR10 models, respectively. Using FP Resnet44 and its quantized versions as targets, attacks were transferred from FP and quantized versions of Resnet20 and Resnet32. Similarly, Mnist C and its quantized versions were used as target models and attacks were transferred from FP and quantized versions of Mnist A and Mnist B. Figure \ref{fig:transferability_res_resnet20_diffcap} shows the results for CIFAR10 models. Similar results were obtained for MNIST and are not shown for the sake of conciseness.

Similarly, to study how model architecture affects transferability, an 11-layer CNN (named as Cifar A for easier reference) was trained on the CIFAR10 dataset. Table~\ref{tbl:Models_cifar_high_cap_arch} shows test accuracy of the model and its quantized versions. Adversarial examples crafted on FP and quantized versions of Resnet20 was then transferred to FP and quantized versions of Cifar A. Figure \ref{fig:transferability_res_resnet20_diffarch} shows the results.

Comparing transferability results in Figure \ref{fig:transferability_res_resnet20} with those in Figure \ref{fig:transferability_res_resnet20_diffcap} and Figure \ref{fig:transferability_res_resnet20_diffarch}, it can be seen that, for the corresponding attack types, the average transferability when the source and targets have different capacity and architecture follows the same pattern as when the source and targets are similar in architecture and capacity. For instance, for the UAP attack in Figure \ref{fig:transferability_res_resnet20} where both source and target networks are different bitwidth versions of Resnet20, the average transferability (for $\xi$ = 0.1) is highest when the attack source is 1-bit Resnet20, while it is lowest when the attack source is a FP Resnet20. This pattern can be seen for the UAP attack in Figure \ref{fig:transferability_res_resnet20_diffcap} as well where the attack sources are Resnet20 and its quantized versions but the target models are Resnet44 and its quantized versions. Moreover, similar observations can be made in Figure \ref{fig:transferability_res_resnet20_diffarch}. The pattern is followed for all attack sources and not just for the sources with the highest or lowest overall adversarial transferability and is true for all attack types. There are very few exceptions, for instance, in the case of JSMA; however these are rare for both MNIST and CIFAR10 models. 

Further, it can be noted from Figure  \ref{fig:transferability_res_resnet20_diffcap} and \ref{fig:transferability_res_resnet20_diffarch} that the transferability is not better when source and target networks are of same architecture as when they are different in architecture. This is inline with the observation in \cite{delving_trans} but is in contrast to the one made in \cite{understand_enhance}.

\noindent
%\paragraph{}\label{summary1} \vspace{-9pt}
\textbf{Summary} 
% \begin{itemize}
% \item \label{exp2:summary_1} 

\emph{The average transferability of an attack among the different bitwidth versions of the same model can be an indication of transferability when the attack is transferred to another model which may be different in not only bitwidth but also in architecture and model capacity.} This suggests that better overall transferability of an attack source when the network architecture and capacity of source and target are similar can also mean better transferability when both source and target differ in architecture and model capacity.

% \item \label{exp1:summary_2} \emph{It may not be necessary for similar architecture networks to have better transferability.} 

% \end{itemize} 

\section{Discussion} \label{sec:discussion}
Based on the results from the experiments, the research questions put forward at the beginning of the study (Section \ref{intro:RQ}) can be answered as follows:

\begin{enumerate}
\item \textbf{RQ1}: What causes some algorithms to have better or worse transferability across variably quantized systems?

It is known from prior works that the transferability of gradient-based attacks among networks of different bitwidths is poor \cite{impact_low_bit, to_compress, uap_transfer}. The transferability experiments with FGSM and UAP showed similar behaviour where both were observed to have poor transferability. However, it was found that the adversarial examples generated using UAP algorithm remained recognizable at higher magnitudes of distortion as compared to FGSM. This resulted in UAP having better transferability than FGSM as the algorithm allowed setting distortion norm to higher values. This suggests that the efficiency of an attack may be improved by enclosing it within the UAP algorithm and increasing the magnitude of distortion.

Furthermore, it was found that the quantization shift phenomena also hinders the transferability of attacks like JSMA that tend to modify individual features to produce adversarial samples. 

\item \textbf{RQ2}: How do model-related properties like architecture and capacity affect transferability among quantized networks?

Although the transferability is poor, it was observed that the average transferability of different attack sources followed the similar pattern as the average transferability when the attacks were transferred among networks which were different bitwidth versions of the same network. This indicates that the transferability of an attack to a black-box model of unknown architecture and capacity can be approximated based on how the attack performs when it is transferred among different bitwidth versions of the source model.

\end{enumerate}

Additionally, algorithms like the Boundary Attack that depend on heuristic search for finding adversarial examples may be highly effective on the source network but may have poor transferability because of the direction of the perturbation not aligning with the adversarial direction. 

%Moreover, transferability was reduced across all attack types, suggesting that quantization does offer some degree of resistance against transfer-based attacks.

% Optimization enables deployment of DNN models on embedded devices, however the security implications must be taken into consideration when applying an optimization technique. Quantization can provide some robustness against transfer-based attacks; nonetheless, being able to estimate the performance of an attack can be considered as a vulnerability.

\section{Conclusion} \label{sec:conclusion}

This paper explores the conditions and properties that can influence the transferability of adversarial examples among networks of varying bitwidths. To this end, transferability analysis on quantized and full-precision networks trained on MNIST and CIFAR10 datasets was performed by considering various algorithms to craft the adversarial examples and also by varying the model-related properties like the architecture and capacity between the source and target networks. Based on a literature review of the current state of the art, this is the first study to consider model-related properties during attack transfers in the context of quantized networks.

% The results show that misalignment of loss gradients and quantization shift can cause poor attack transferability. Further, when attacks are transferred to a black-box model with unknown capacity, architecture, and bitwidth, transferability may be poor but is predictable based on the average transferability of the attack across different bitwidth models derived from same source model. 

Optimization enables deployment of DNN models on embedded devices, however the security implications must be taken into consideration when applying an optimization technique. Quantization can provide some robustness against transfer-based attacks; nonetheless, being able to estimate the performance of an attack can be considered as a vulnerability.

\begin{acks} 
    This work has been funded by https://ki-lok.itpower.de/ (German Federal Ministry for Economic Affairs and Climate Action, project no.: 19121007A) and https://iml4e.org (German Federal Ministry for Education and Research, project no.: 01IS21021C)
\end{acks}

\bibliographystyle{ACM-Reference-Format}
\bibliography{bibliography}

\appendix

\section{Model details} \label{app:model_details}

The model architectures are based on MNIST and CIFAR models in Tensorpack repository \cite{tensorpack}. Specifically, the models are as defined in the following scripts: 
\begin{itemize}
\item Mnist A is based on \url{https://github.com/tensorpack/tensorpack/blob/master/examples/basics/mnist-convnet.py}
\item Resnet20 is based on \url{https://github.com/tensorpack/tensorpack/blob/master/examples/ResNet/cifar10-resnet.py}
\item Cifar A is based on \url{https://github.com/tensorpack/tensorpack/blob/master/examples/basics/cifar-convnet.py}. 
\end{itemize}

Some adjustments in image augmentations, optimizer, and learning rates were made. The training hyperparameters are mentioned as follows:

\subsection{Mnist A} \label{sub:mnistA_arch}

Training hyperparameters are as below:
\begin{itemize}
\item\emph{Normalization:}  Features are normalized to [0,1]

\item \emph{Optimizer:} Adam optimizer with learning rate($\alpha$)=0.001, $\beta_1$=0.9, $\beta_2$=0.999, $\varepsilon$=1e-8  

\item \emph{Regularization:}  Weight decay on all FC layers  

Regularization hyperparameter ($\lambda$): 1e-5  

\item \emph{Loss:}  Cross-entropy

\end{itemize}

\subsection{ResNet20} \label{sub:resnet20_arch}

Training details are as below:

\begin{itemize}
\item \emph{Normalization:} Features are normalized to [0,1]

\item \emph{Optimizer:} Momentum optimizer with $\gamma$ = 0.9

\item \emph{Regularization:} Weight decay on all layers

Regularization hyperparameter ($\lambda$): 2e-4

\item \emph{Loss:} Cross-entropy

\item \emph{Learning Rate:}

Scheduled to change as (epoch, value): (1, 0.1), (32, 0.01), (48, 0.001), (72, 0.0002), (82, 0.00002)
\end{itemize}

\section{Adversarial Examples from Various Models} \label{app:adv_example_sample_images}

During the experiments, hyperparameters as mentioned in Table \ref{tbl:mnisthyperparam} were used to craft adversarial samples. The selection of these attack hyperparameters depended on whether they produced recognizable images (although distorted). 

Adversarial examples were created by selecting 1000 random samples that were classified correctly by both source and target networks. Performance of the networks was then measured against the corresponding adversarial samples. The top-5 adversarial samples from these 1000 samples generated using each hyperparameter setting for all attack types are presented in the subsequent Figures.

\begin{figure*}[htpb]
\centering
%---FGSM_eps_025------
\begin{subfigure}{.3\textwidth}
  %\centering
  \includegraphics[width=.6\linewidth]{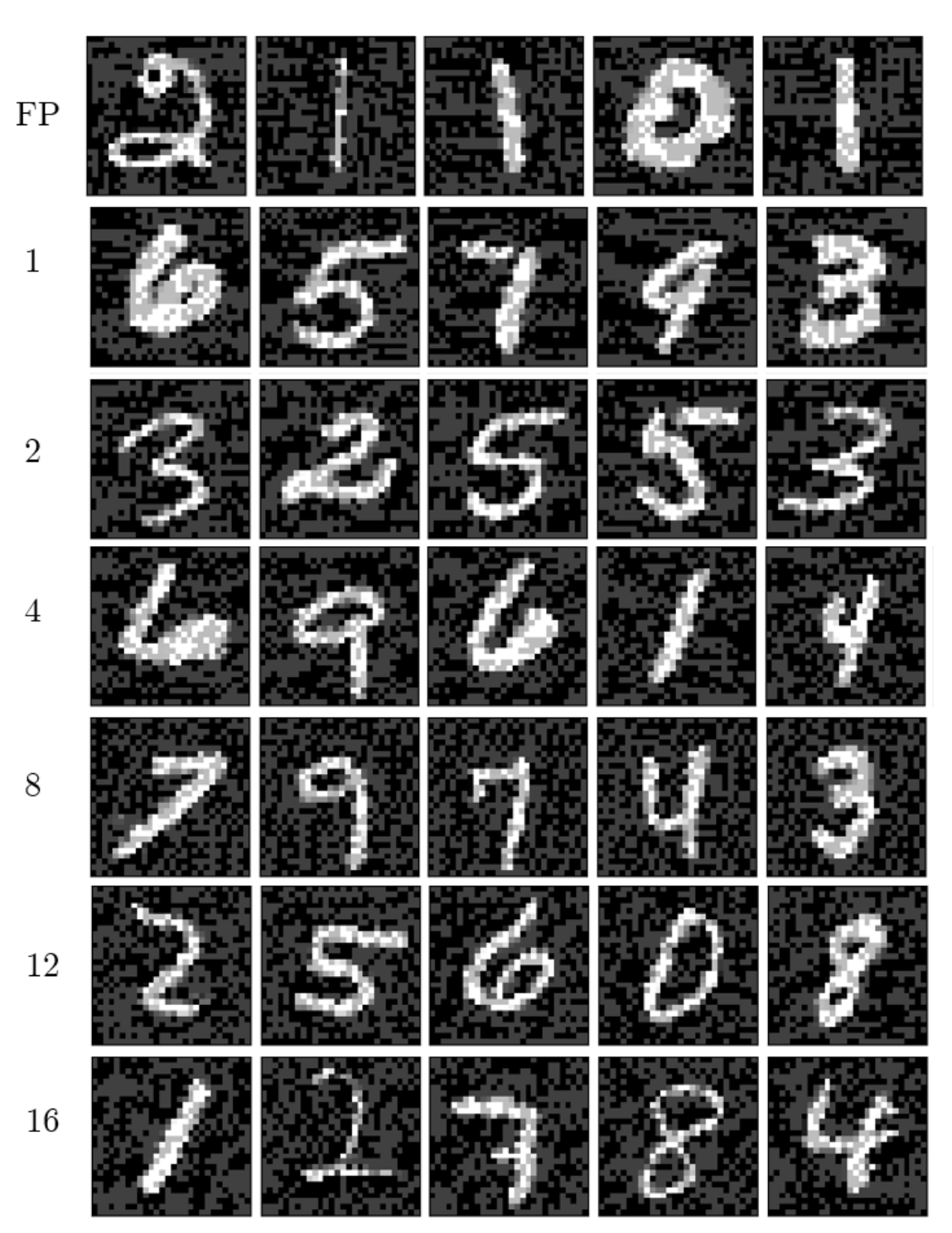}
  \captionsetup{justification=raggedright,singlelinecheck=false}
  \caption*{FGSM ($\varepsilon = 0.25$)}
\end{subfigure}%
%---JSMA_1_01------
\begin{subfigure}{.3\textwidth}
  %\centering
  \includegraphics[width=.6\linewidth]{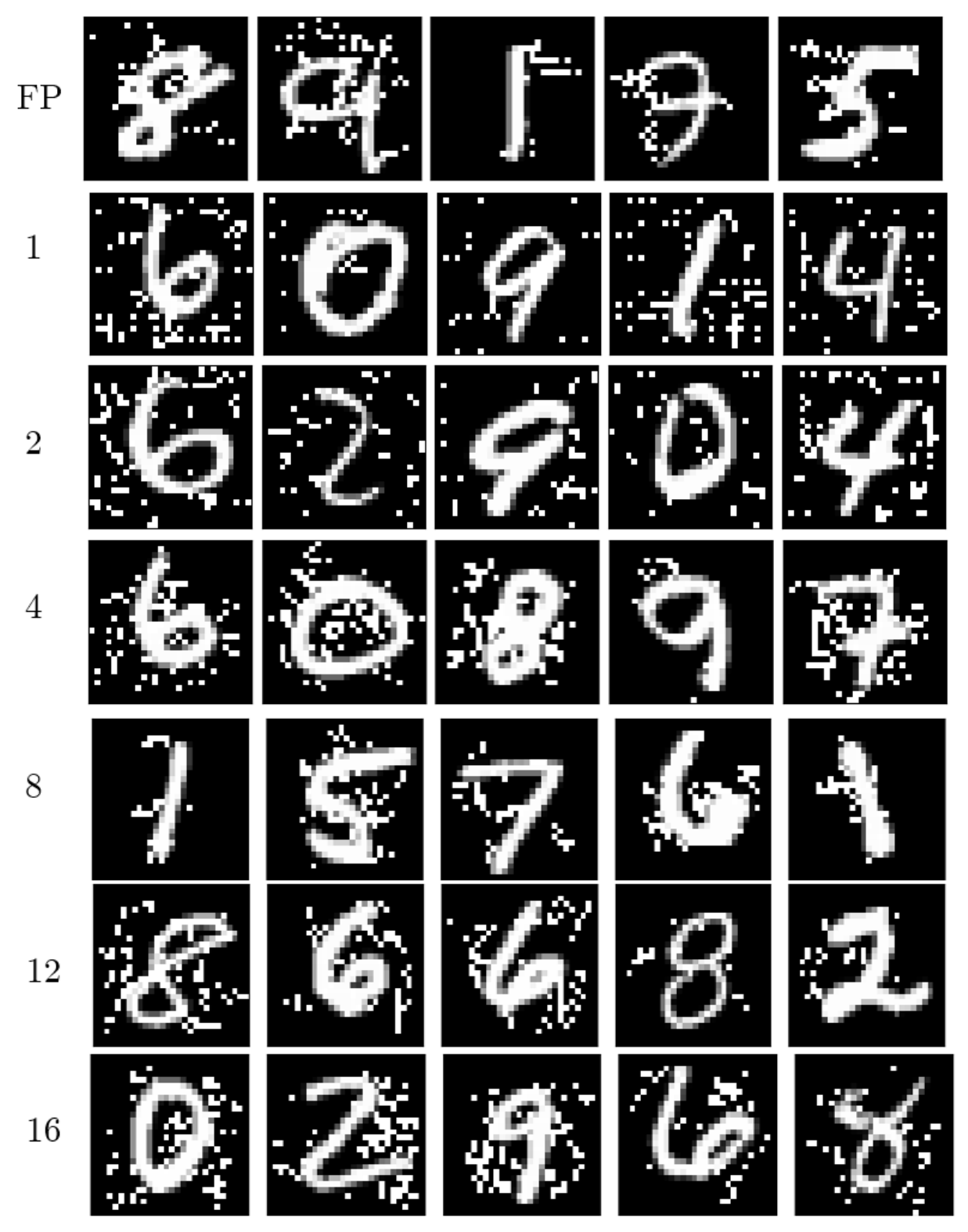}
  \captionsetup{justification=raggedright,singlelinecheck=false}
  \caption*{JSMA ($\theta = 1$, $\gamma = 10\%$)}
\end{subfigure}%
%---BA_15_ITT------
\begin{subfigure}{.3\textwidth}
  %\centering
  \includegraphics[width=.6\linewidth]{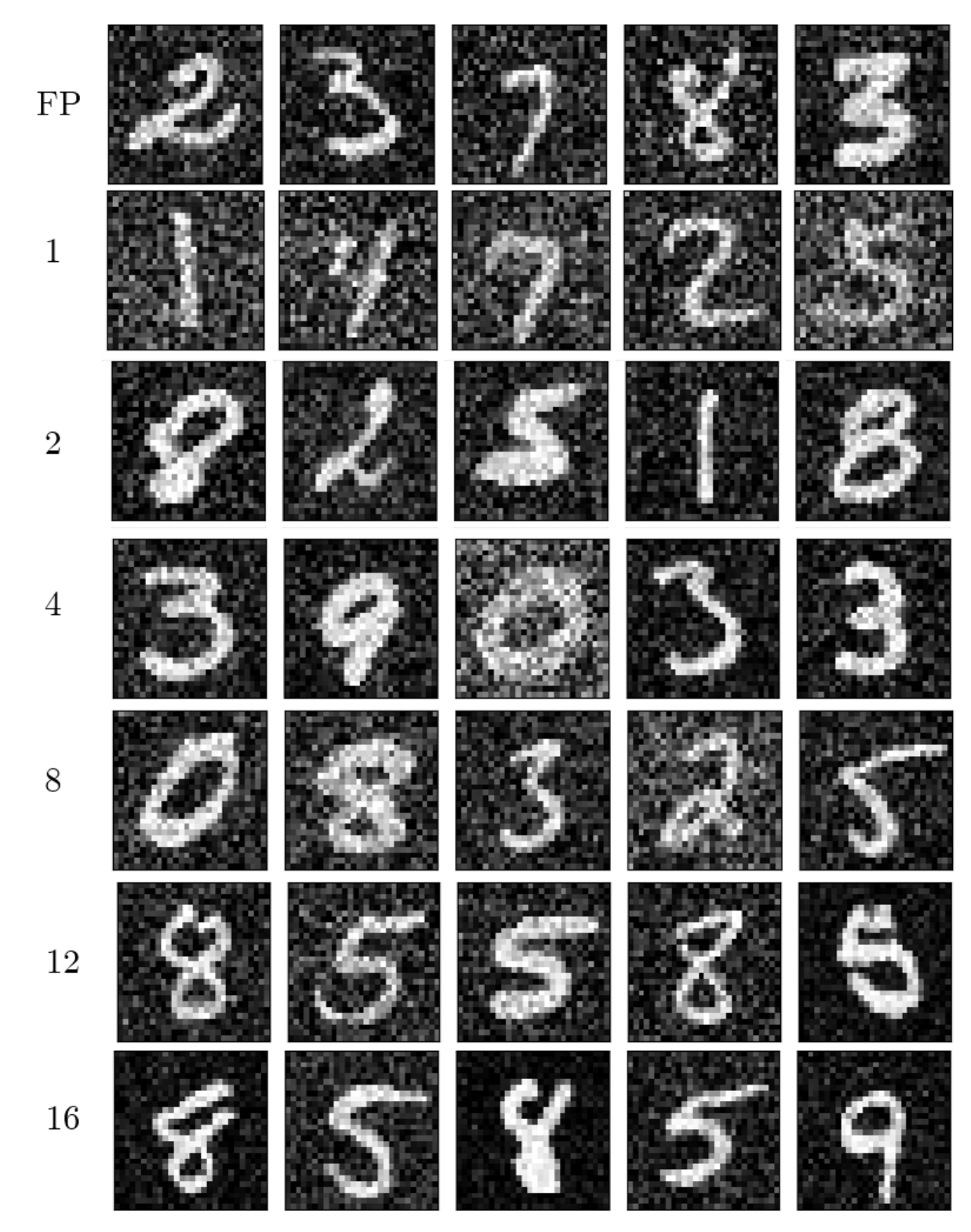}
  \captionsetup{justification=raggedright,singlelinecheck=false}
  \caption*{BA (\emph{i} = 15)}
\end{subfigure}

%---UAP_eps_01_xi_06------
\begin{subfigure}{.3\textwidth}
  %\centering
  \includegraphics[width=.6\linewidth]{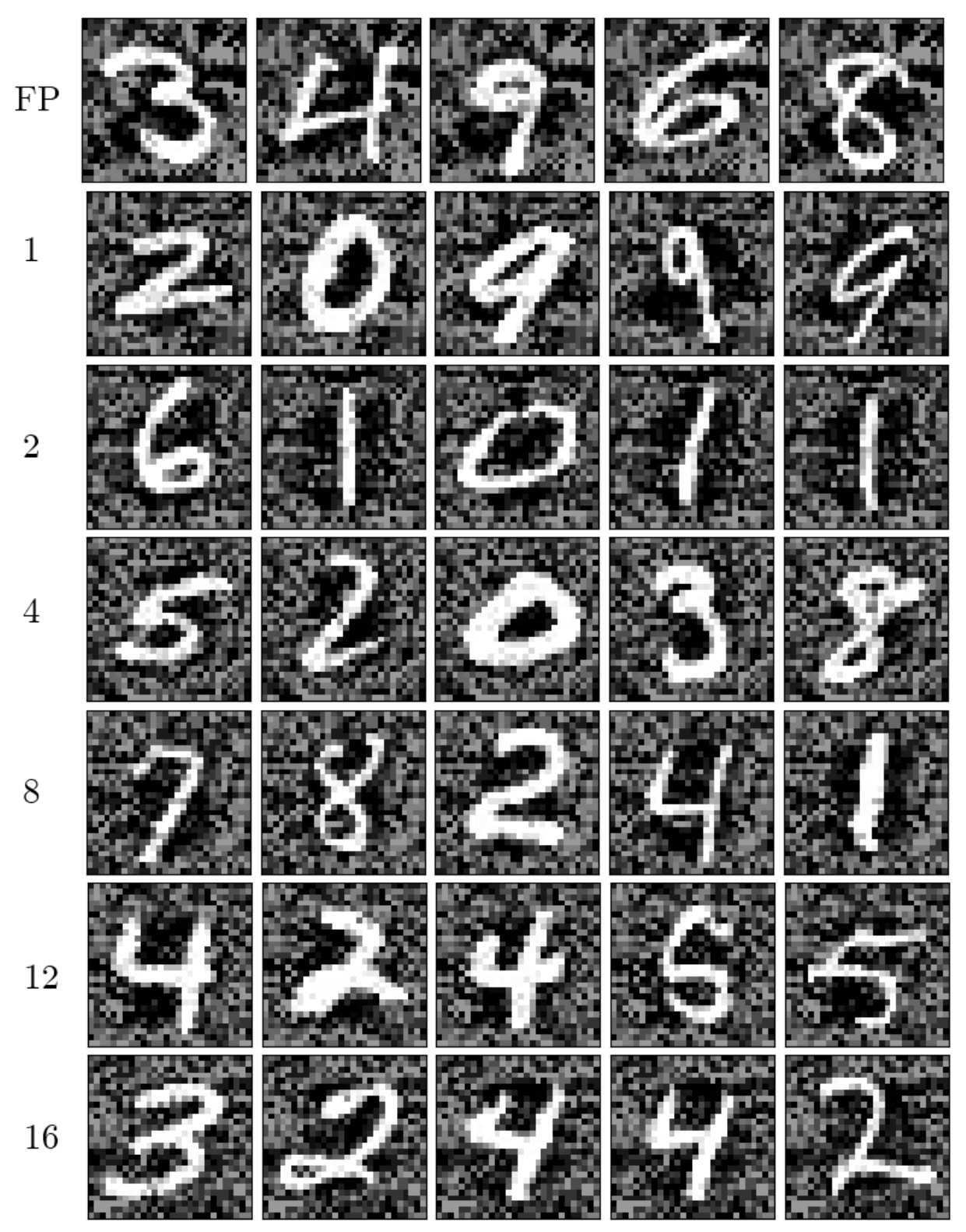}
  \captionsetup{singlelinecheck=false}
  \caption*{UAP ($\varepsilon = 0.1$, $\xi = 0.6$)}
\end{subfigure}%
%---CW_BSS_20_K_5_ITT_30-------
\begin{subfigure}{.3\textwidth}
  %\centering
  \includegraphics[width=.6\linewidth]{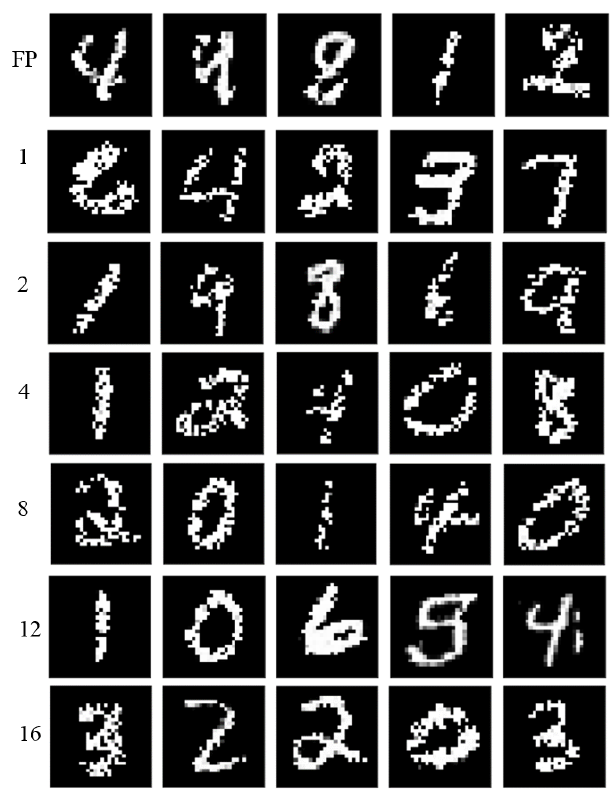}
  \captionsetup{singlelinecheck=false}
  \caption*{CW ($\kappa = 5$, $i = 25$, $b_s = 20$)}
\end{subfigure}
%-----------------------------
\captionsetup{justification=raggedright,singlelinecheck=false}
\caption[Adversarial samples created using FGSM, JSMA, UAP, and BA on various bitwidths of Mnist A model.]{Adversarial samples created using FGSM, JSMA, BA, UAP, and CW attack on various bitwidths of Mnist A model.}
\Description{Adversarial samples created using FGSM, JSMA, BA, UAP, and CW attack on various bitwidths of Mnist A model.}
\label{fig:mnist_fgsm_images}
\end{figure*}

\begin{figure*}[htpb]
\centering
%---FGSM_eps_005------
\begin{subfigure}{.3\textwidth}
  %\centering
  \includegraphics[width=.6\linewidth]{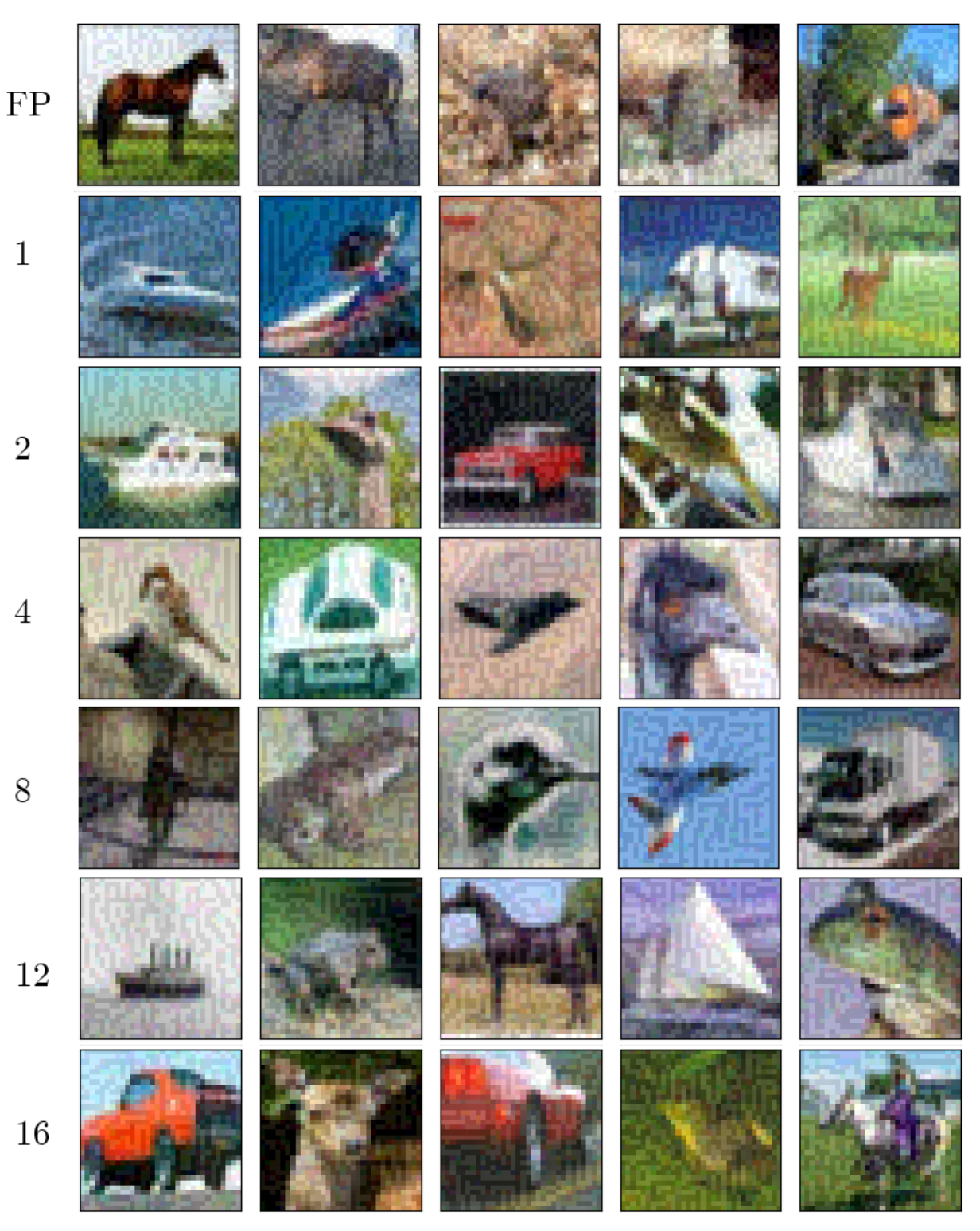}
  \captionsetup{justification=raggedright,singlelinecheck=false}
  \caption*{FGSM ($\varepsilon = 0.05$)}
\end{subfigure}%
%---JSMA_03_005------
\begin{subfigure}{.3\textwidth}
  %\centering
  \includegraphics[width=.6\linewidth]{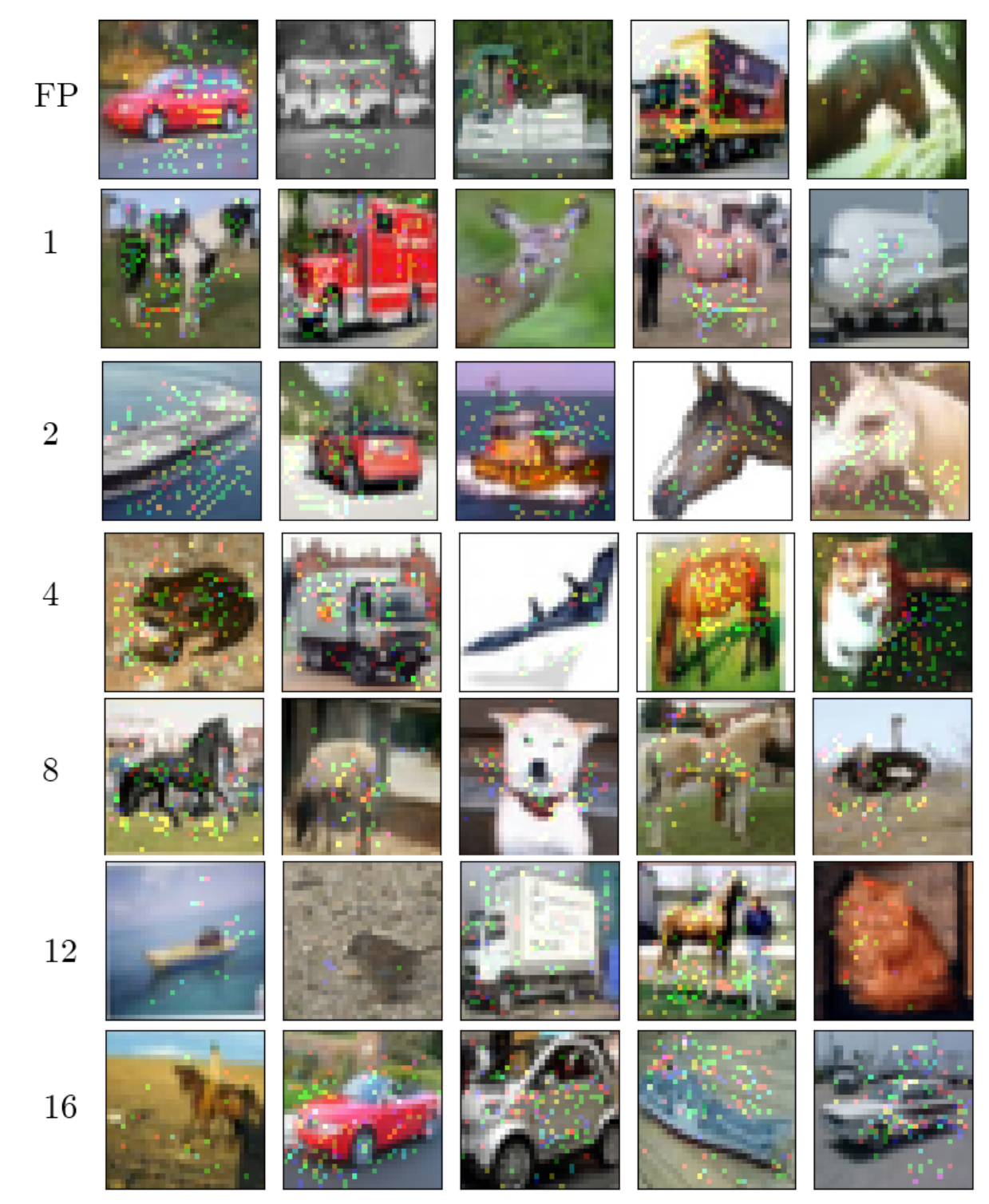}
  \captionsetup{justification=raggedright,singlelinecheck=false}
  \caption*{JSMA ($\theta = 0.3$, $\gamma = 5\%$)}
\end{subfigure}%
%---BA_12_ITT------
\begin{subfigure}{.3\textwidth}
  %\centering
  \includegraphics[width=.6\linewidth]{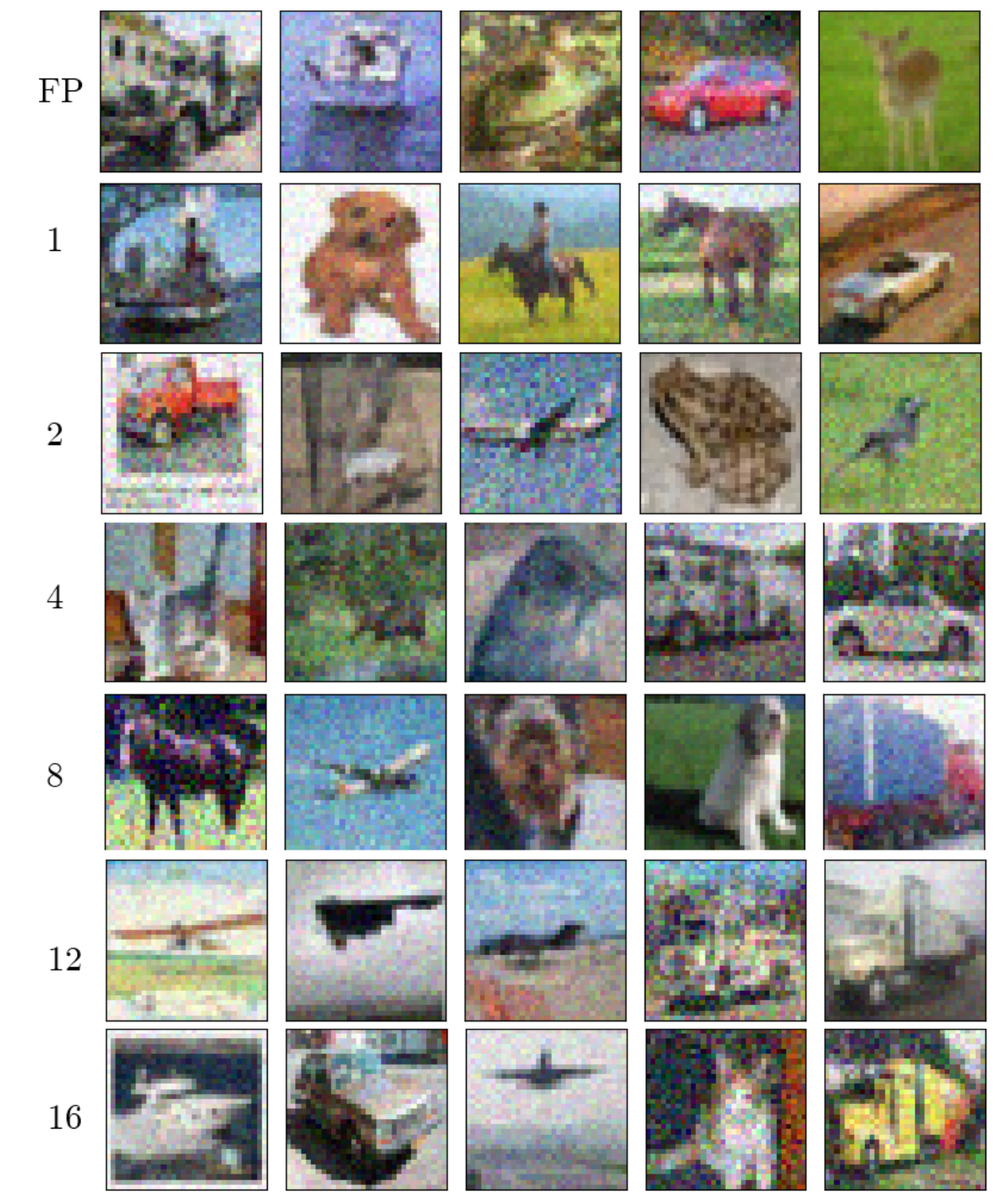}
  \captionsetup{justification=raggedright,singlelinecheck=false}
  \caption*{BA (\emph{i} = 12)}
\end{subfigure}

%---UAP_eps_001_xi_01------
\begin{subfigure}{.3\textwidth}
  %\centering
  \includegraphics[width=.6\linewidth]{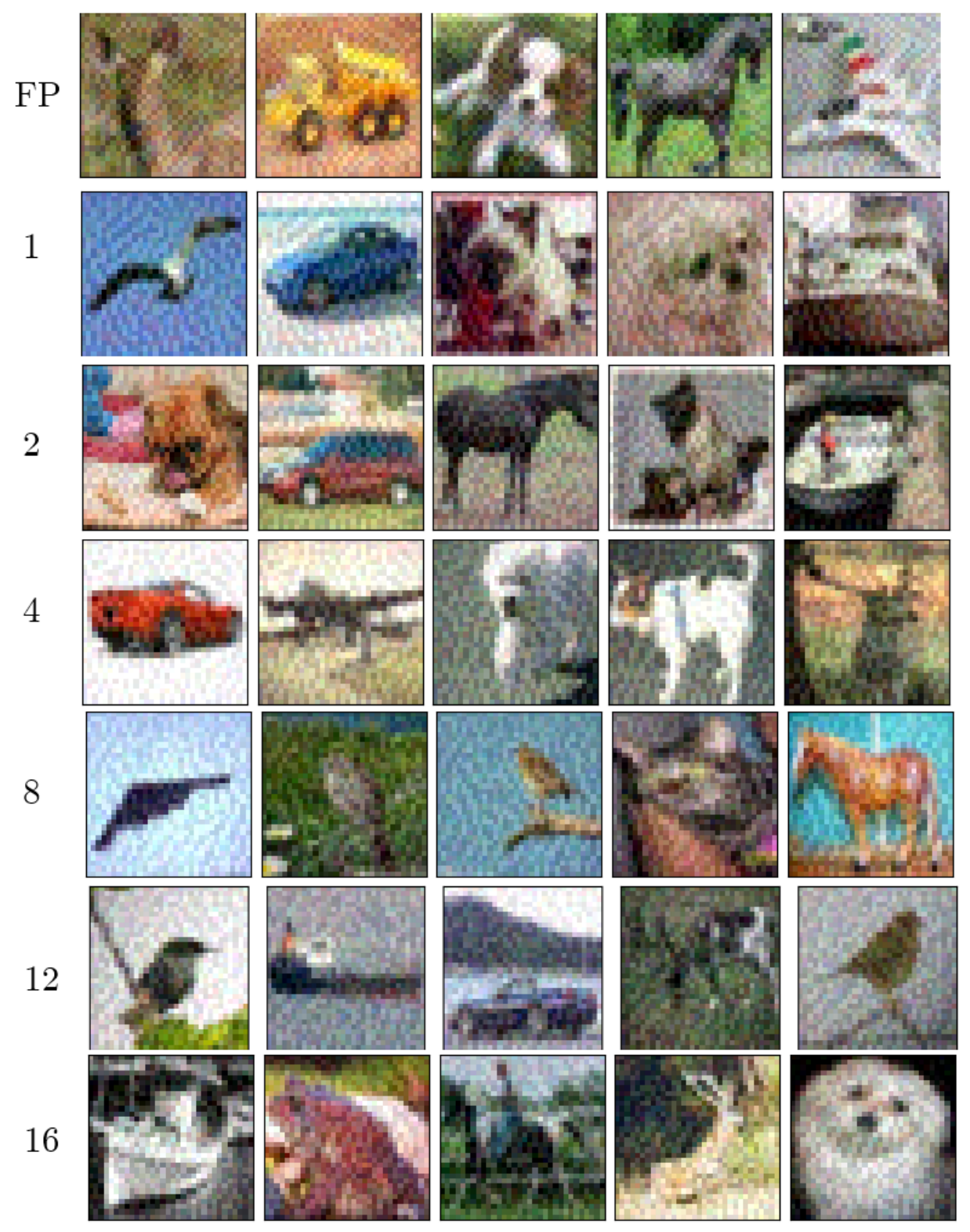}
  \captionsetup{singlelinecheck=false}
  \caption*{UAP ($\varepsilon = 0.01$, $\xi = 0.1$)}
\end{subfigure}%
%---CW_BSS_20_K_5_ITT_30-------
\begin{subfigure}{.3\textwidth}
  %\centering
  \includegraphics[width=.6\linewidth]{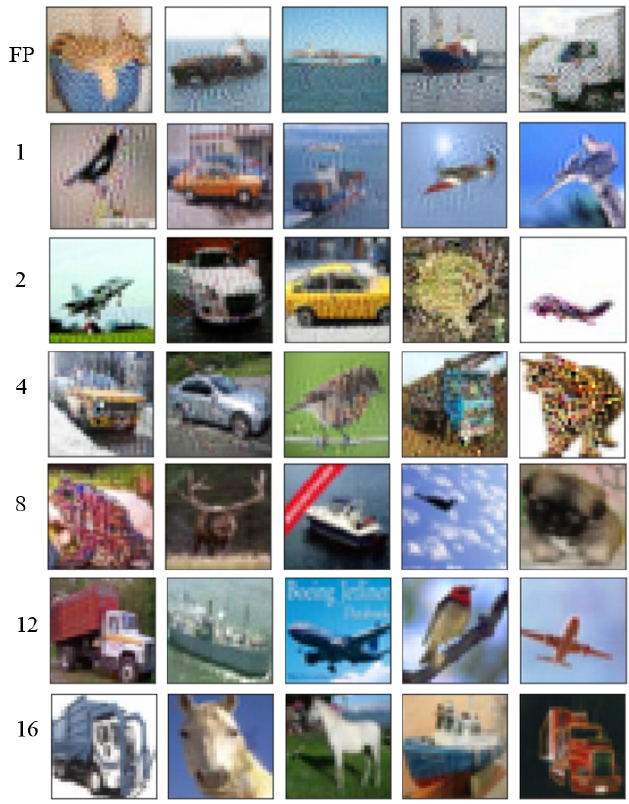}
  \captionsetup{singlelinecheck=false}
  \caption*{CW ($\kappa = 5$, $i = 25$, $b_s = 20$)}
\end{subfigure}

%-----------------------------
\captionsetup{justification=raggedright,singlelinecheck=false}
\caption[Adversarial samples created using on various bitwidths of Resnet20 model.]{Adversarial samples created using FGSM, JSMA, BA, UAP, and CW attack on various bitwidths of Resnet20 model.}
\Description{Adversarial samples created using FGSM, JSMA, BA, UAP, and CW attack on various bitwidths of Resnet20 model.}
\label{fig:cifar_fgsm_images}
\end{figure*}

Figure \ref{fig:mnist_fgsm_images} and \ref{fig:cifar_fgsm_images}  depict adversarial examples crafted on different bitwidths of the Mnist A and Resnet20 model, respectively. The labels on the left of each set of 5 images represent the bitwidth of the network on which the samples were created.

%------------------------------------------------------MNIST---IMAGES
%hp = put figure "here" in a page for "only float"

%------------------------------------------------------CIFAR10---IMAGES---FGSM
%hp = put figure "here" in a page for "only float"

\end{document}